\documentclass{article}

    \PassOptionsToPackage{numbers, compress}{natbib}

\usepackage{subcaption}
\usepackage{multirow}
\usepackage[utf8]{inputenc}
\usepackage{csquotes}
\usepackage{graphicx}
\usepackage{tabularx}
\usepackage{mathtools}
\usepackage{tabulary}
\usepackage{booktabs}       %
\usepackage{nicefrac}       %
\usepackage{microtype}      %
\usepackage{booktabs}
\usepackage{pifont}
\usepackage{amsmath}

\usepackage{color}

\usepackage{epsfig}

\usepackage{caption}

\usepackage{svg}
\usepackage{adjustbox}
\usepackage{longtable}
\usepackage{xspace}

\usepackage{enumitem} %
\usepackage{tcolorbox}
\usepackage{xcolor}
\definecolor{tolblue}{rgb}{0,0.08,0.45}

\definecolor{lightyellow}{rgb}{1, 0.95, 0.85}

\definecolor{darkred}{rgb}{0.85, 0.3, 0.2}  %
\definecolor{mydarkblue}{rgb}{0,0.08,0.45}

\usepackage[utf8]{inputenc} %
\usepackage[T1]{fontenc}    %
\usepackage[pagebackref=true,colorlinks=true,linkcolor=darkred,citecolor=cyan,urlcolor=darkred]{hyperref}\usepackage{url}            %
\usepackage{booktabs}       %
\usepackage{amsfonts}       %
\usepackage{nicefrac}       %
\usepackage{microtype}      %
\usepackage{xcolor}         %
\usepackage{cleveref}
\crefname{section}{Sec.}{Secs.}
\crefname{appendix}{App.}{Apps.}
\crefname{figure}{Fig.}{Figs.}

\usepackage{tikz}
\usepackage{comment}
\usepackage{amsmath,amssymb} %
\usepackage{color}

\newcommand{\cmnt}[1]{}

\usepackage{xspace}

\newcommand{\llava}{LLaVA-1.5\xspace}

\newcommand{\alphasubnet}{$\alpha$-SubNet\xspace}

    \usepackage[final]{neurips_2024}

\usepackage{authblk}
\usepackage[utf8]{inputenc} %
\usepackage[T1]{fontenc}    %
\usepackage{hyperref}       %
\usepackage{url}            %
\usepackage{booktabs}       %
\usepackage{amsfonts}       %
\usepackage{nicefrac}       %
\usepackage{microtype}      %
\usepackage{xcolor}         %

\title{
Skipping Computations in Multimodal LLMs
}

\author{
  \textbf{Mustafa Shukor}$^{1}$ \thanks{Contact: \{firstname.lastname\}@sorbonne-universite.fr} \quad \quad \quad \quad \quad \textbf{Matthieu Cord}$^{1,2}$  \\ 
  \makebox[0pt][c]{\textnormal{$^{1}$Sorbonne University,} \textnormal{$^{2}$Valeo.ai}} \vspace{-0.5cm}
}

\begin{document}

\maketitle

\begin{abstract}

Large Language Models (LLMs) have demonstrated remarkable success in both textual and multimodal domains. However, this success often comes with substantial computational costs, particularly when handling lengthy sequences of multimodal inputs. This has sparked many efforts focusing on enhancing efficiency during training and inference. In this study, we investigate the computation redundancy in Multimodal Large Language Models (MLLMs) during inference. We propose different methods to skip computations, such as skipping entire blocks, FFN or self-attention (SA) layers. Additionally, we explore parallelizing certain layers, such as FFN and SA layers. Our findings validate that (1) significant amount of computations can be avoided at inference time, especially for tasks such as Visual Question Answering (VQA). (2) Skipping computations during training can recover 97\% of the original performance, even when skipping half of the blocks or removing 70\% of the weights. Alternatively, (3) properly training with smaller LLMs can yield comparable performance to LLMs 2 or 3 times larger. To conclude, we extend our investigation to recent MLLMs, such as LLaVA-1.5, showing similar observations. Our work show that there is redundant computations inside MLLMs and thus the potential for significantly improving inference costs without sacrificing performance. The code is available here: \href{https://github.com/mshukor/ima-lmms}{https://github.com/mshukor/ima-lmms}.

\end{abstract}

\section{Introduction}

Large Language Models (LLMs) \cite{hoffmann2022trainingchinchilla,zhang2022opt,chowdhery2022palm,openai2023gpt,touvron2023llamav2} have been a major step towards human level intelligence. These models are capable of achieving reasonable scores on almost any textual task that can be done by humans.

Beyond the textual realm, LLMs are now the main building block for large multimodal models (LMMs) or multimodal LLMs (MLLMs) \cite{alayrac2022flamingo,chen2023palix,driess2023palme,openai2023gpt}. However, training LLMs on more modalities requires significantly more computation resources due to the complexity of mulitmodal inputs. Multimodal inputs incur longer sequence length, additional encoders to tokenize different modalities and additional latency to preprocess each example.

Recent approaches have tried to overcome this burden by freezing all pretrained model parameters and training
only the mapping module \cite{merullo2022linearlylimber,shukor2023epalm,manas2022mapl,liu2023visualllava,dai2023instructblip,yang2022zerofrozenbilm,depalm}. These models only train a modest number of parameters (few millions) and are able to attain reasonable performance on a wide range multimodal tasks. Besides the trainable parameters, complementary works tackle also data-efficiency and avoid costly multimodal pretraining \cite{shukor2023epalm,manas2022mapl,depalm}. When parameter and data-efficiency are combined, the training cost is reduced significantly, and becomes affordable by consumer grade GPUs.

\begin{figure*}[h]
    \centering
    \includegraphics[width=0.95\textwidth]{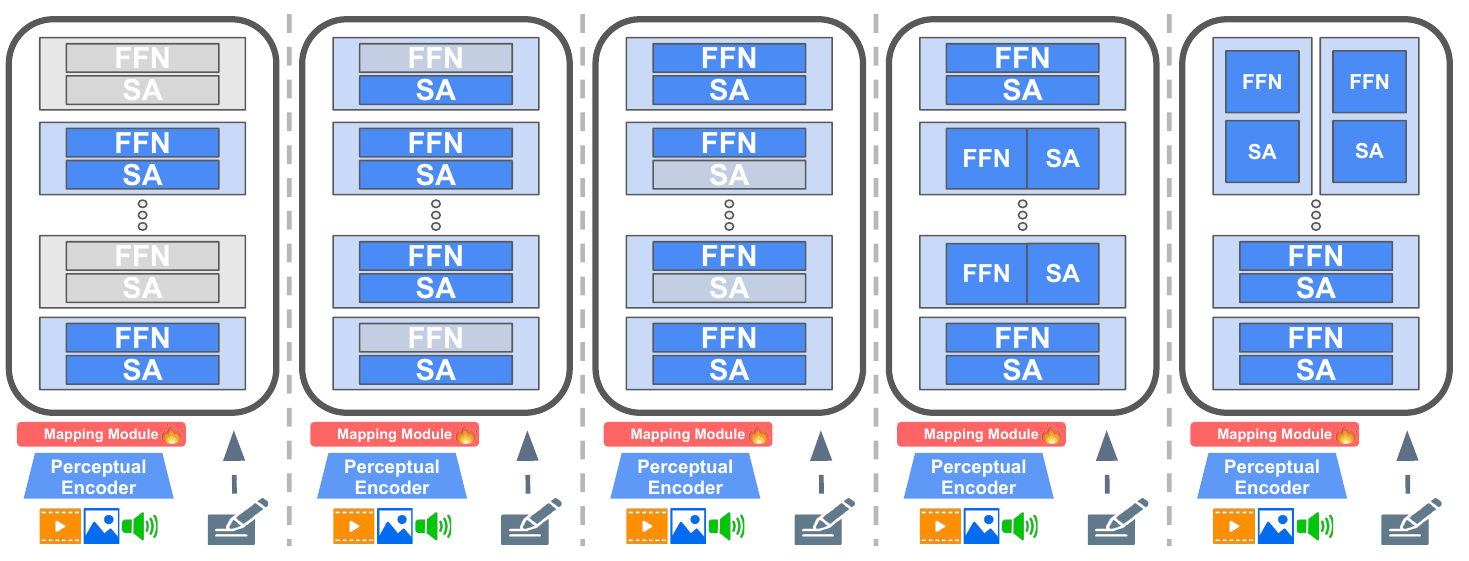}
    \caption{\footnotesize \textbf{Illustration of the proposed techniques to skip and parrallelize computations in multimodal LLMs.} From left to right: Skipping entire blocks (Skip Block), skipping only the FFN layers (Skip FFN), skipping the self-attention layers (Skip SA), parrallelizing the FFN and SA, parrellizing entire blocks. These are applied each interval I of layers, starting at a specific layer (sl).}
\label{fig:iou}
\end{figure*}

However, reducing the inference cost of MLLMs is a problem that gather little of attention. These models, bottelnecked with an LLM, are significantly slow at inference and consume large amount of RAM and storage memory, hindering their deployment in the real world.

In this work, we hypothesize that LLMs are highly overparametrized for general multimodal tasks, and contain redundant parameters, layers and blocks that can be bypassed.

Our work is also inspired by the recent study \cite{shukor2024implicit} that highlights the slowly changing embeddings for both textual and perceptual tokens, a phenomenon that has been observed with LLMs \cite{song2024sleb,gromov2024unreasonable,liu2023dejavu}.

We investigate the computation redundancy at different granularity levels. We focus on skipping computations for both visual and textual tokens during autoregressive generation. 
Specifically, we start our study with pretrained models and test if we can skip entire blocks, FFN or SA layers and individual neurons, without any additional training. In addition, we also experiment with parrallizing the SA and FFN layers, or 2 entire blocks. 

We continue our investigation focusing on skipping entire blocks, but during training. This helps to reduce both the training and inference cost. We show that we can retain more than 97\% of the performance when training the mapping module with highly sparse LLMs; such as skipping half of the layers or removing more than 70\% of the parameters. Finally, we show that almost the same original performance can be retained by properly training multimodel models with smaller LLMs.

In a nutshell, we propose a framework to study and compare different task-agnostic compression methods, for image, video and audio language tasks. Our study led to the following findings:
\begin{itemize}
    \item Skipping computations for the generated textual tokens leads only to slight performance degradation, especially for VQA tasks.
    \item Training the mapping module while skipping computations can retain almost the original performance, even when 70\% of the parameters are removed or 50\% of the blocks are skipped.
    \item Properly training with smaller LLMs (\emph{e.g.} OPT-2.7B) can achieve the same performance as larger ones (\emph{e.g.} OPT-6.7B).
    \item Similar observations holds for the larger-scale MLLMs such as \llava.
\end{itemize}
    
\section{Related work}

\paragraph{LLMs and LMMs.} Large Language Models (LLMs) \cite{brown2020languagegpt3,hoffmann2022trainingchinchilla,chowdhery2022palm,scao2022bloom,zhang2022opt,touvron2023llamav2,openai2023gpt} serve as the foundational components for contemporary Large Multimodal Models (LMMs). Most of these LMMs either utilize frozen LLMs \cite{alayrac2022flamingo,awadalla2023openflamingo,laurencon2023obelics} or fine-tune LLMs post-initialization \cite{chen2022pali,chen2023palix,driess2023palme}. Beyond scale, LLMs have facilitated the development of unified models capable of addressing multiple multimodal tasks \cite{shukor2023unival,mizrahi20234m,lu2023unified,wang2022unifyingofa,lu2022unifiedio,diao2023writedavinci} in a unified framework. Efficient adaptation of unimodal models \cite{merullo2022linearlylimber,shukor2023epalm,yang2022zerofrozenbilm,manas2022mapl,llamaadapter,koh2023groundingfromage,zhu2023minigpt,dai2023instructblip,moon2023anymal} represents a vital research avenue aiming to circumvent costly training by maintaining LLMs frozen and training a handful of adaptation parameters. Despite being significantly more efficient, they also compete with end-to-end trained models \cite{li2021alignalbef,vicha,meter,li2022blip,singh2022flava} across image/video/audio and language tasks \cite{shukor2023epalm,wang2023gpt4video,wu2023nextgpt,panagopoulou2023xinstrcutblip}. However, despite their training efficiency, these models still incur significant costs at inference.

\paragraph{Compression for LLMs.} Model compression has long been a research focus, particularly with the rise of large models. Approaches have been developed to effectively and scalably compress LLMs with hundreds of billions of parameters \cite{brown2020languagegpt3,zhang2022opt,scao2022bloom}. Post-training pruning methods aim to sparsify the model using a limited number of examples without additional model training. These methods employ layer-wise optimization \cite{frantar2022optimal,frantar2022spdy,kwon2022a,frantar2023sparsegpt} or achieve efficiency with just a few model inferences \cite{sun2023wanda}. While most pruning approaches target unstructured pruning, structured pruning results in actual wall clock time reduction \cite{ma2023llmpruner,liu2023dejavu}. Post-training quantization \cite{chee2024quip,shao2023omniquant,frantar2023optq,egiazarian2024extreme,yvinec2024rex} reduces the precision of model parameters to 8 bits, 4 bits, or lower, leading to increased efficiency and support across a wide range of hardware architectures. Early exiting methods directly generate the output from intermediate layers without using the last LLM layers \cite{schuster2022confident,chen2023ee}. While conditional computations approaches, skip computations based on the input sample \cite{ainslie2023colt5,shazeer2016outrageouslymoe,raposo2024mixturemod}, our approach is static and input-agnostic.

\paragraph{Compression for multimodal models.} While numerous approaches focus on compressing LLMs, fewer methods have been proposed for multimodal models. Existing approaches include knowledge distillation from powerful models \cite{fang2021compressing,wu2023tinyclip}, unstructured pruning based on the Lottery Ticket Hypothesis (LTH) \cite{gan2022playinglth,tan2022end}, or structured pruning \cite{pmlr-v202-shi23eupop}. However, most of these methods require a significant amount of additional cost when applied to very large models such as MLLMs.

\section{Framework for compressing preceptually augmented LLMs}

\subsection{General MLLMs framework} 

We investigate a general architecture comprising a frozen LLM, a trainable mapping module ($C$), and frozen perceptual encoders ($E_M$) for various modalities (M), such as image (I), video (V), and audio (A). The input to the LLM, denoted as X, is a concatenation of textual tokens ($T=[t_1, ...,t_{n_T}]$) and multimodal or perceptual tokens, referred to as the prompt ($P=[p_1, ...,p_{n_P}]$). The prompt is generated by encoding the modality-specific input (XM) with the corresponding encoder and projecting it to the LLM input space using the mapping module. This can be expressed as follows:
\begin{align}
    O = LLM\left(X\right) \quad \quad \quad \quad \quad \quad   \nonumber \\
    X = [P; T] \quad P = C(E_M(XM)) \quad T = E_T(XT) 
\end{align}

The LLM consists of N blocks ${B^l}_{l \in \{0, ..., N-1\}}$, where each block slightly refine the input tokens $X^l$:

\begin{align}
\label{eq:block}
    X^{N} = \sum_{l=0}^{N-1}{B^l(X^l)},
\end{align}

And each block B can be expressed as:

\begin{align}
\label{eq:in_block}
    X^{l+1} = X_1 + FC2(g(FC1(LN2(X_1)))) \nonumber \\
  X_1 = X^{l} + SA(LN1(X^{l})), \quad \quad 
\end{align}

We mainly focus on the parameter/data-efficient setup, where both the LLMs and the perceptual encoders are kept frozen and only a light-weight mapping module (few millions parameters) is trained.

\subsection{Efficient MLLMs baselines}  
We follow the same setup of previous works  \cite{shukor2023epalm,manas2022mapl,liang2022modularpromptfuse,depalm}, where we finetune one mapping module for each multimodal task. We train multiple models on image, video, and audio-text datasets. We adopt the baselines proposed in \cite{shukor2024implicit}, that adopt a lightweight transformer with learnable queries and self-attention mechanisms to attend to perceptual tokens. This transformer operates in a low-dimensional space, facilitated by down/up projection layers, and limits the number of learnable queries. Our baselines are similar to \cite{depalm}, however, we use significantly less number of learnable query (\emph{e.g.}, 10) and prioritize a deeper architecture consisting of 5 blocks with hidden dimension of 256 over a wider one with a hidden dimension. We found this deeper architecture to work better in practice with reduced number of parameters. These baselines are trained with OPT-6.7B \cite{zhang2022opt}, and our study is complemented with Vicuna-v1.5-7B \cite{zheng2024judgingvicunav15}. We utilize different powerful encoders for images (CLIP \cite{radford2021learning}), videos (X-CLIP \cite{ma2022xclip}), and audios (AST \cite{gong21b_interspeech_ast}).

\subsection{Implementation details}
We adopt the implementation details outlined in \cite{shukor2024implicit}. Specifically, we employ the AdamW optimizer with a learning rate of 2e-4, which decreases using a cosine annealing scheduler to a minimum of 1e-5. During training, we use a total batch size of 16 for captioning and 64 for VQA datasets. We set the number of epochs to 20 to ensure convergence, although many models converge within just a couple of epochs. The best checkpoint is selected for evaluation; for instance, the image captioning model typically converges after approximately 4 epochs. Training is conducted on 8 V100 GPUs, and the duration varies depending on the task; for instance, each epoch takes about 30 minutes for the large VQAv2 dataset, while smaller datasets like Audiocaps and MSVD-QA require around 10 minutes per epoch.

\subsection{Datasets and metrics}
Following prior works, we select several public multimodal datasets that encompass two representative tasks: captioning and question-answering (QA) across various modalities, including image (VQAv2 \cite{goyal2017makingvqav2}, COCO caption \cite{lin2014microsoftcoco}), video (MSVD \cite{msvd_msrvtt}, MSRVTT \cite{Xu_2016_CVPR_msrvtt}), and audio (Audiocaps \cite{audiocaps}). For QA datasets, we report accuracy in an open-ended generation setup with exact match, while for captioning, we report the CIDEr metric.

\section{Skipping computations for MLLMs}
\label{sec:skipping_main}

\subsection{Skipping computations}

\paragraph{Method.} We propose to skip entire layers in an input and task-agnostic manner. LLMs consists of many repetitive blocks which we argue that they are redundant and can be bypassed. In addition, previous works have demonstrated the slowly changing embeddings in LLMs \cite{song2024sleb,gromov2024unreasonable,liu2023dejavu} or MLLMs \cite{shukor2024implicit}. Specifically, when skipping entire blocks, \Cref{eq:block} can be written as:
\begin{align}
\label{eq:skip_block}
X^{N-1} &= \sum_{\substack{l \geq sl; \ l \ \% \ I = 0; \ l \in \{0, ..., N-1\}}} B^l(X^l),
\end{align}
Which means that the skipping happens starting from layer $sl$ and each interval $I$ (\emph{e.g.}, $sl=0$ and $I=2$ skip half the blocks). Inside the block, we also investigate if we can skip FFN or SA layers. To skip FFNs, each interval (I) of layers, \Cref{eq:in_block} can be written as: 
\begin{align}
\label{eq:skip_ffn}
 X^{l+1} = X^{l} + SA(LN1(X^{l})),
\end{align}
And similarly when skipping SA layers:
\begin{align}
\label{eq:skip_sa}
    X^{l+1} = X^{l}  + FC2(g(FC1(LN2(LN1(X^l))))
\end{align}

\begin{figure*}[h]
    \centering
    \begin{minipage}{\linewidth}
    \centering
        \begin{minipage}{.19\linewidth}
        \begin{subfigure}[b]{\textwidth}
                \includegraphics[width=1.0\textwidth]{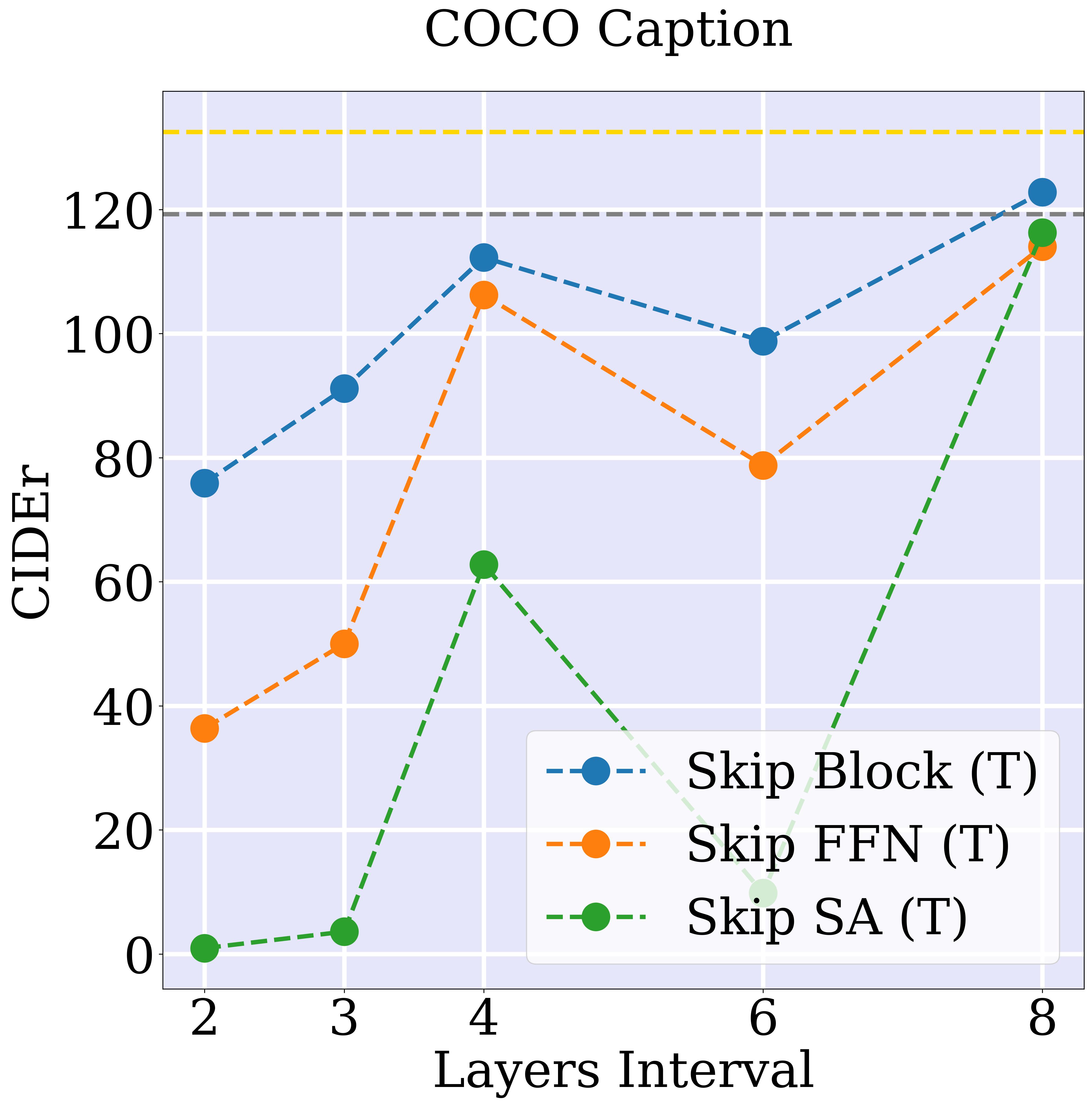}
            \end{subfigure}
        \end{minipage}%
        \begin{minipage}{.19\linewidth}
        \begin{subfigure}[b]{\textwidth}
                \includegraphics[width=1.0\textwidth]{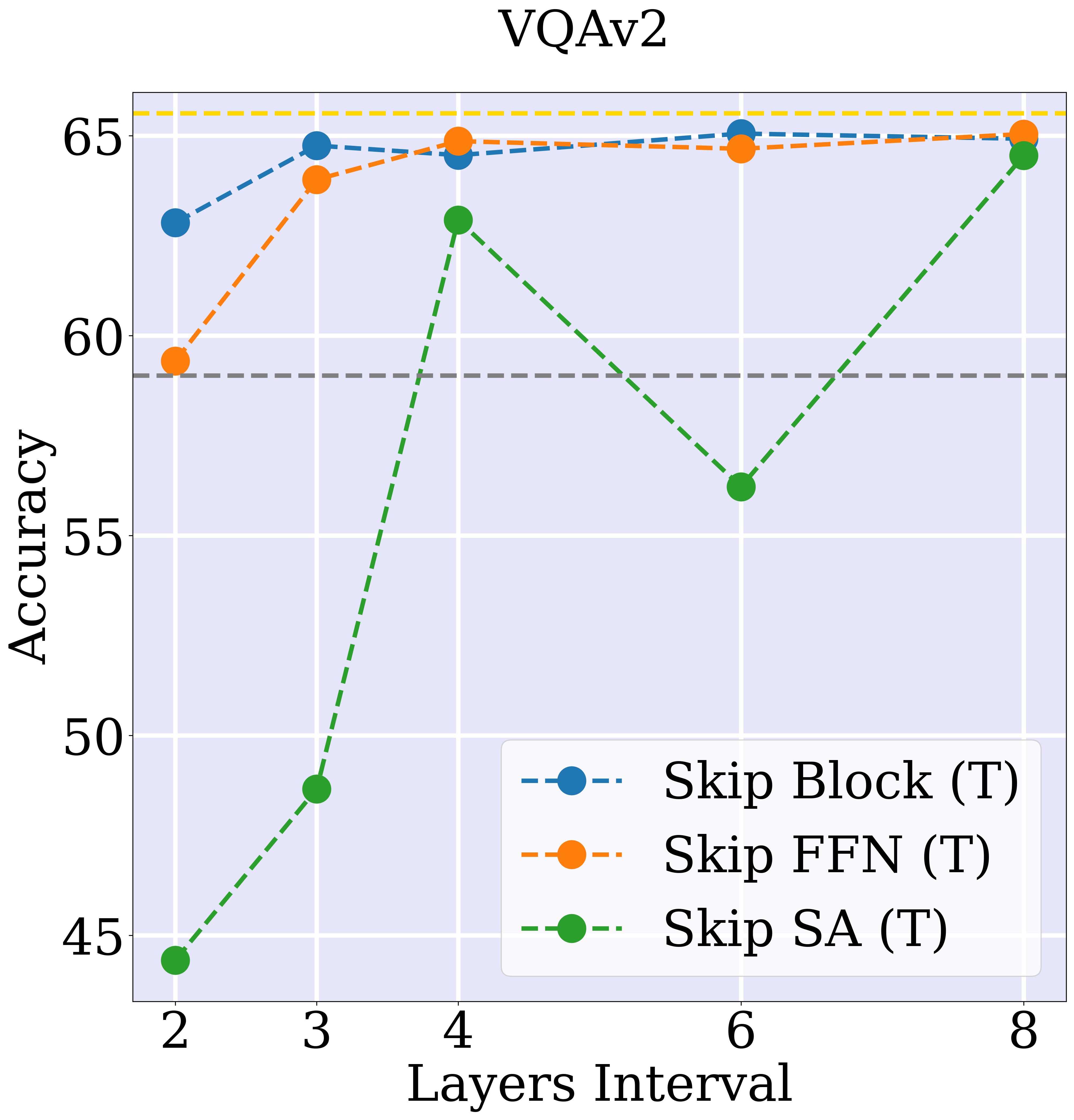}
            \end{subfigure}
        \end{minipage}%
        \begin{minipage}{.19\linewidth}
        \begin{subfigure}[b]{\textwidth}
                \includegraphics[width=1.0\textwidth]{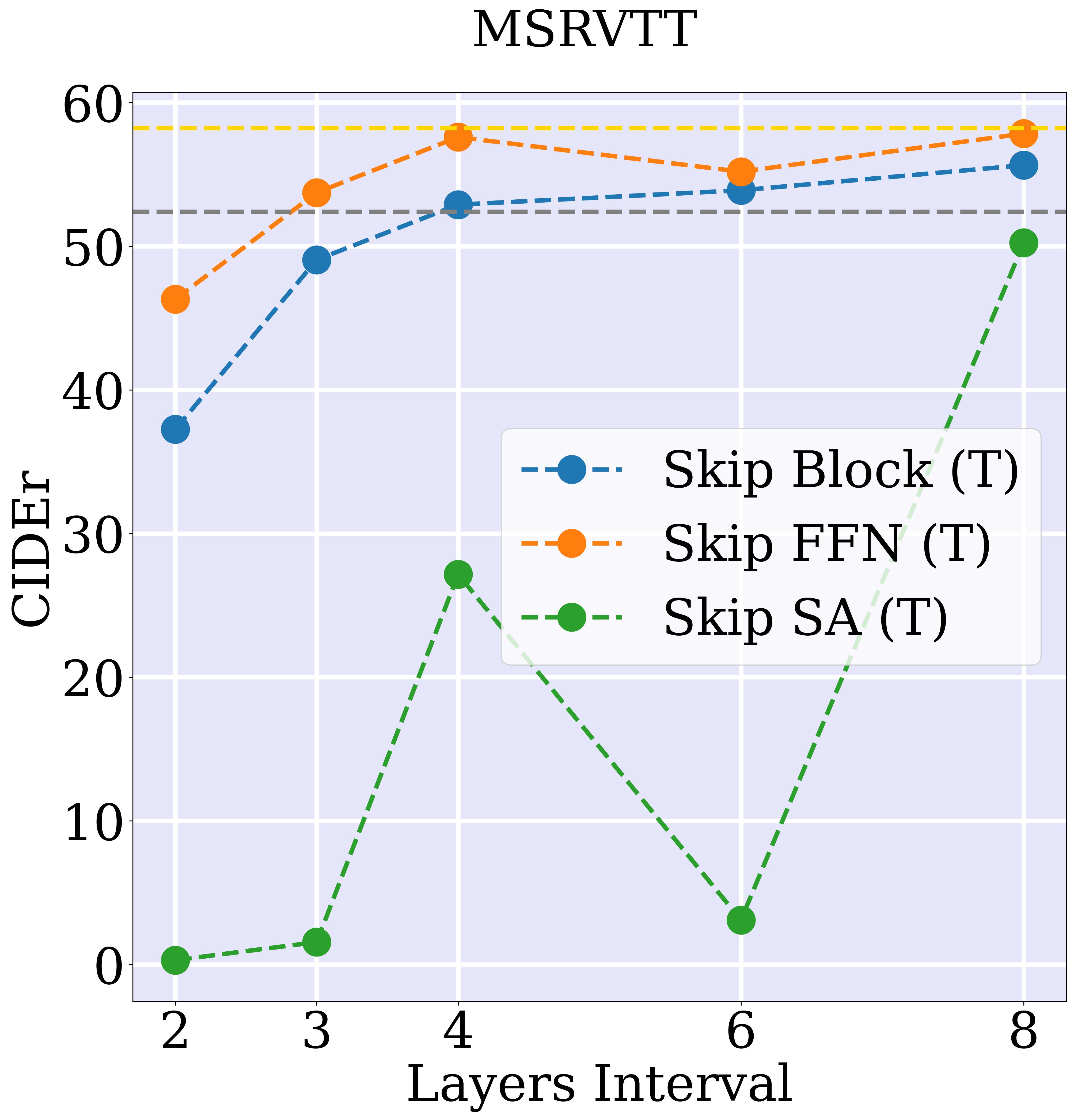}
            \end{subfigure}
        \end{minipage}%
        \begin{minipage}{.19\linewidth}
        \begin{subfigure}[b]{\textwidth}
                \includegraphics[width=1.0\textwidth]{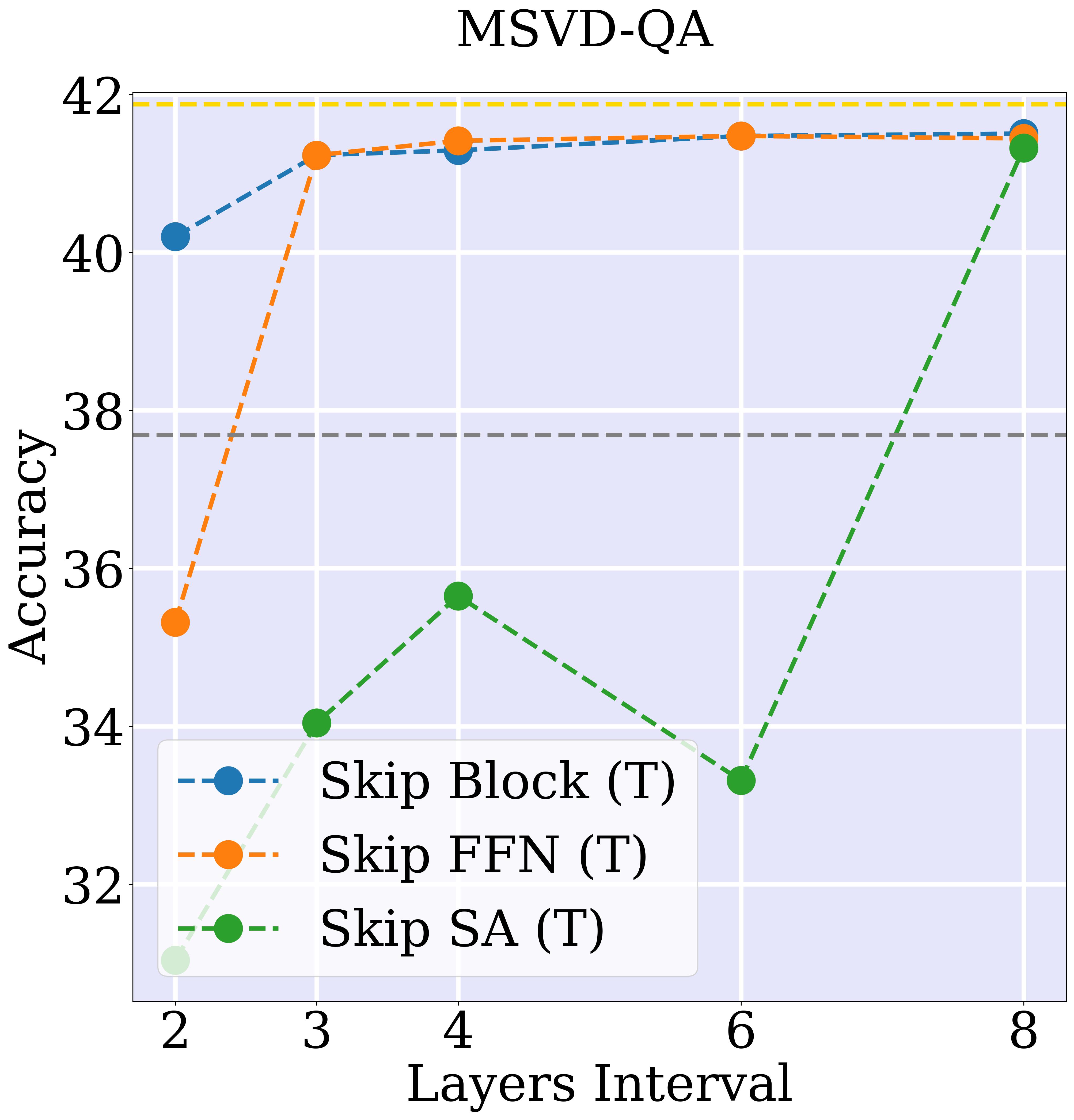}
            \end{subfigure}
        \end{minipage}%
        \begin{minipage}{.19\linewidth}
        \begin{subfigure}[b]{\textwidth}
                \includegraphics[width=1.0\textwidth]{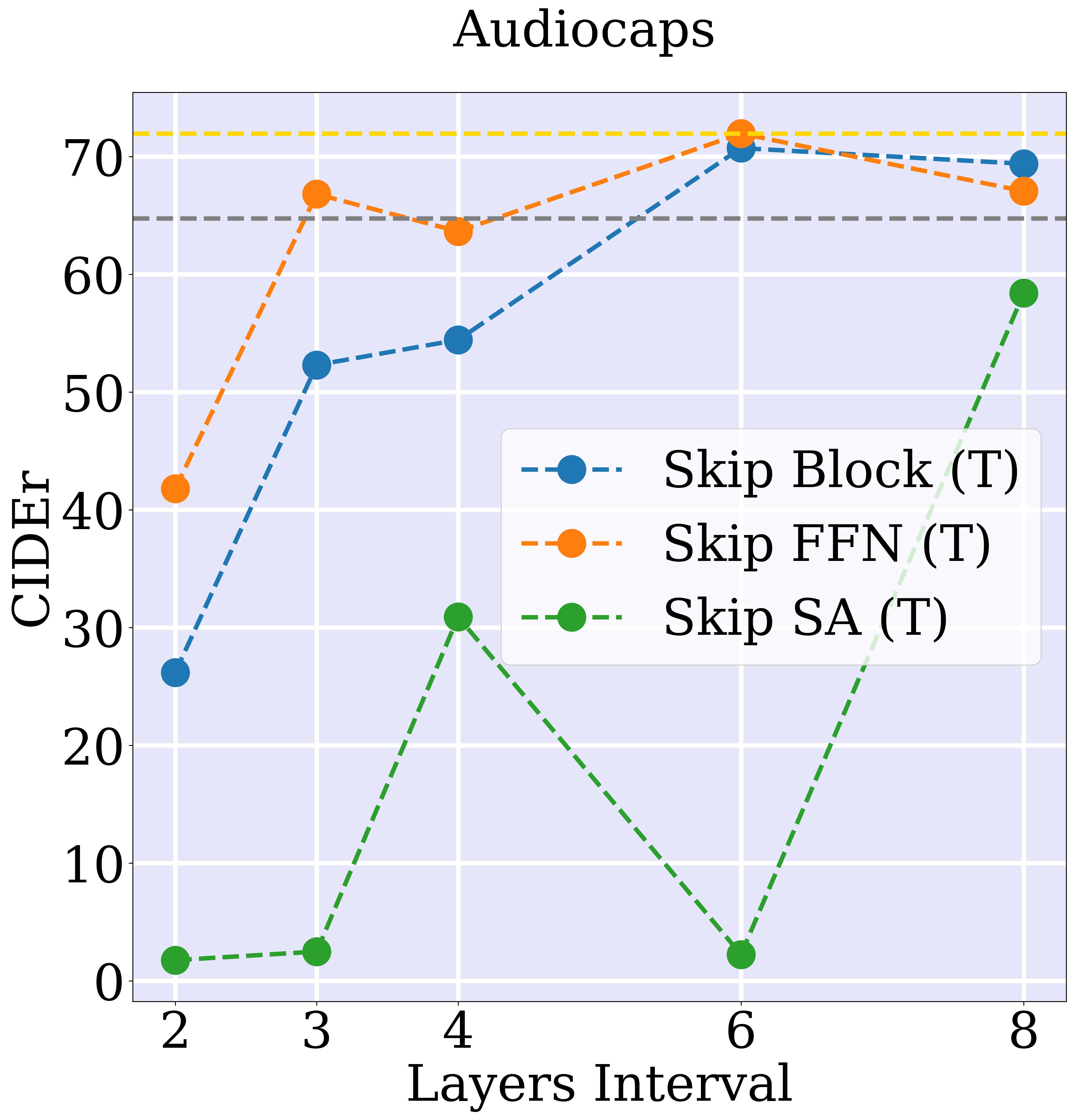}
            \end{subfigure}
        \end{minipage}%
    \end{minipage}%

\caption{\textbf{Skipping computations inside MLLMs.} We skip entire blocks (Skip Block), FFN (Skip FFN) or SA layers (Skip SA). The skipping start at layer 4 and happen each couple of layers (Layer Interval). The gray line indicate 90\% of original performance (shown in yellow).}
\label{fig:skip_opt_main}
\end{figure*}

\paragraph{Experimental results.} \Cref{fig:skip_opt_main} presents a comparison of skipping blocks, feed-forward networks (FFNs), or self-attention (SA) layers across various multimodal datasets. For question-answering (QA) tasks, we observe that we can skip up to 33\% of the blocks while retaining over 90\% of the original performance. However, captioning tasks pose greater challenges due to the larger number of generated textual tokens, with the ability to skip between 15\% and 25\% of the blocks depending on the dataset. In general, skipping entire blocks yields the best results, whereas skipping SA layers results in the lowest performance, underscoring the significance of SA layers for these models.

\begin{figure*}[h]
    \centering
    \begin{minipage}{\linewidth}
    \centering
        \begin{minipage}{.19\linewidth}
        \begin{subfigure}[b]{\textwidth}
                \includegraphics[width=1.0\textwidth]{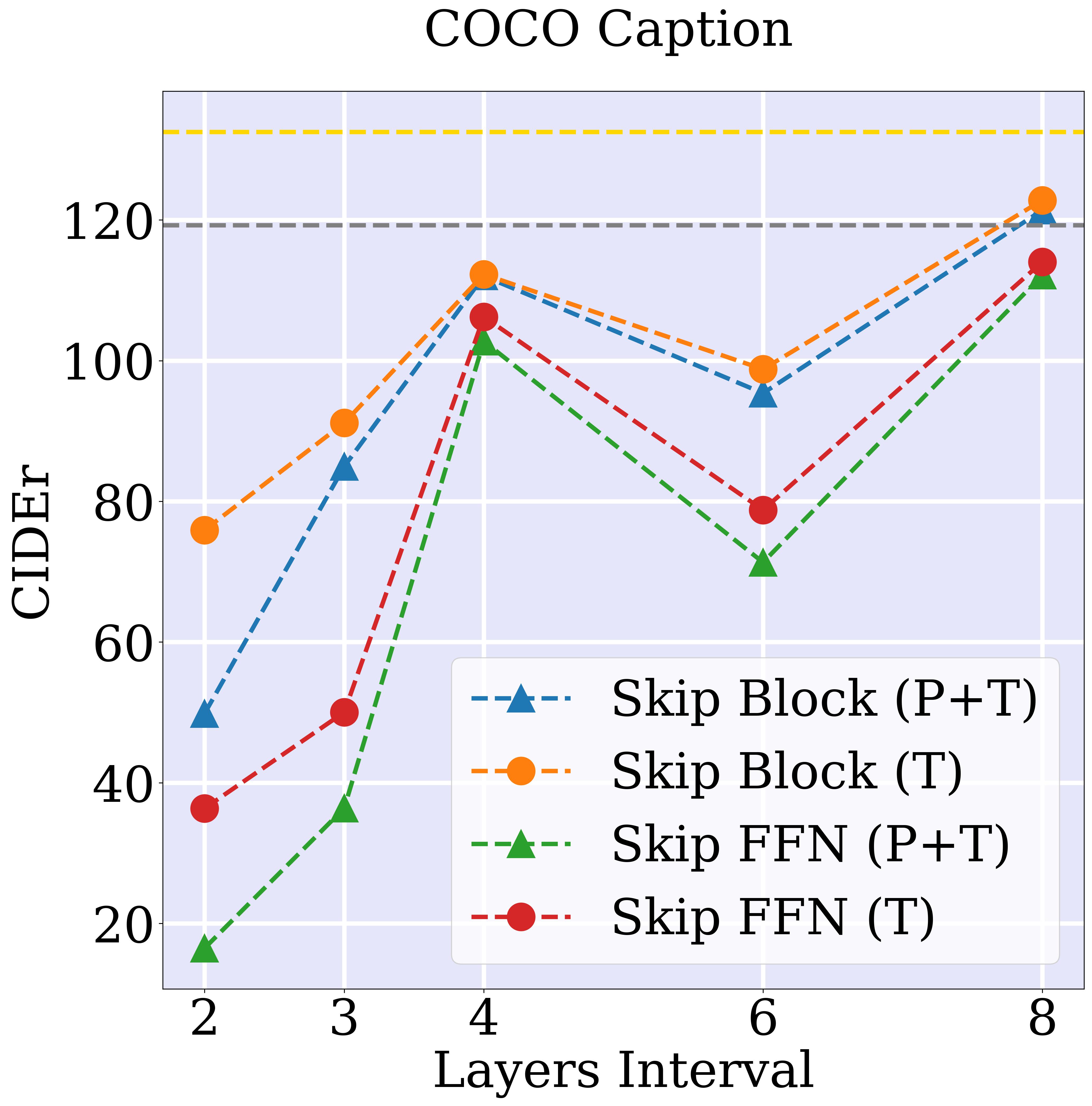}
            \end{subfigure}
        \end{minipage}%
        \begin{minipage}{.19\linewidth}
        \begin{subfigure}[b]{\textwidth}
                \includegraphics[width=1.0\textwidth]{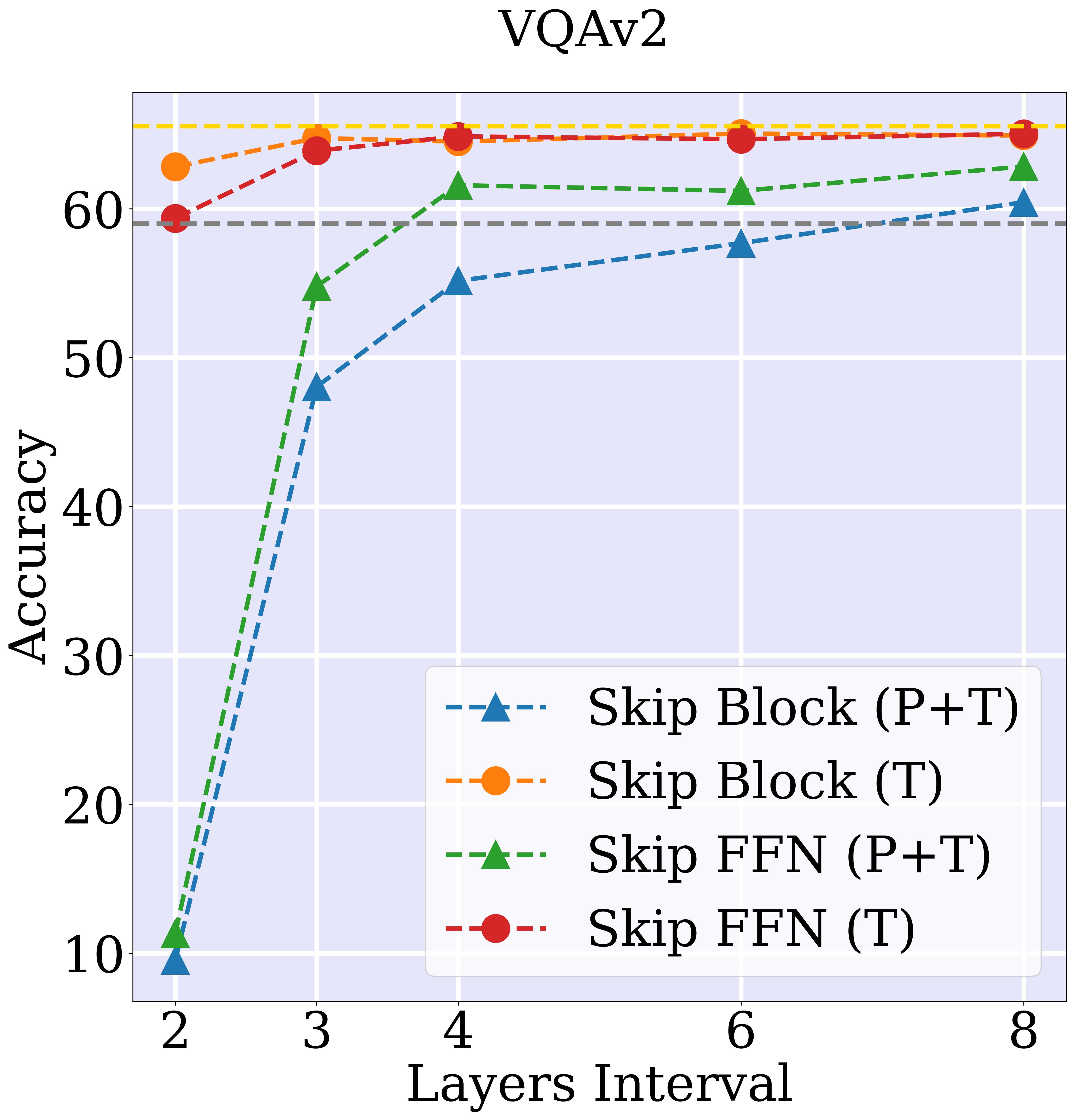}
            \end{subfigure}
        \end{minipage}%
        \begin{minipage}{.19\linewidth}
        \begin{subfigure}[b]{\textwidth}
                \includegraphics[width=1.0\textwidth]{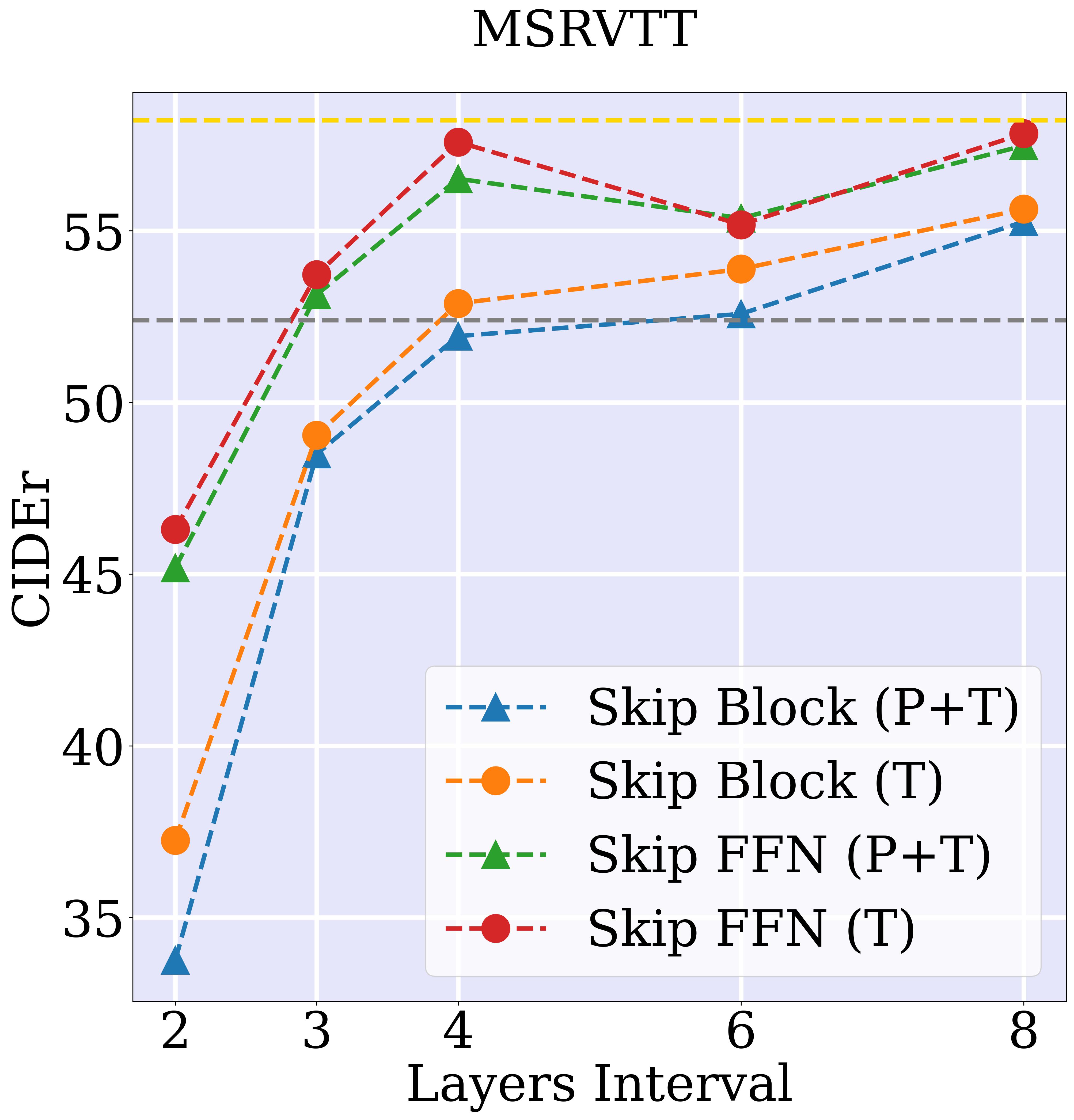}
            \end{subfigure}
        \end{minipage}%
        \begin{minipage}{.19\linewidth}
        \begin{subfigure}[b]{\textwidth}
                \includegraphics[width=1.0\textwidth]{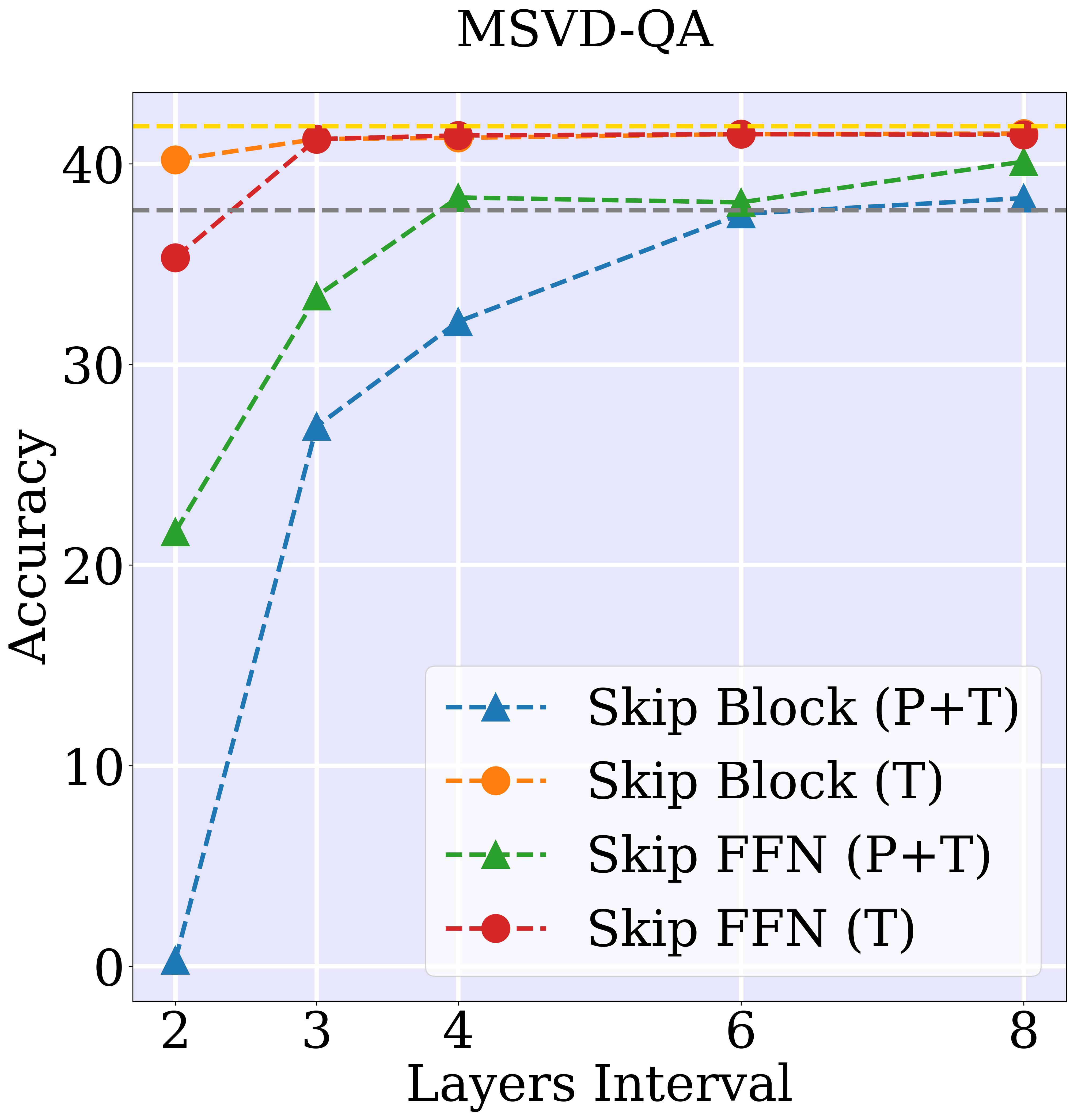}
            \end{subfigure}
        \end{minipage}%
        \begin{minipage}{.19\linewidth}
        \begin{subfigure}[b]{\textwidth}
                \includegraphics[width=1.0\textwidth]{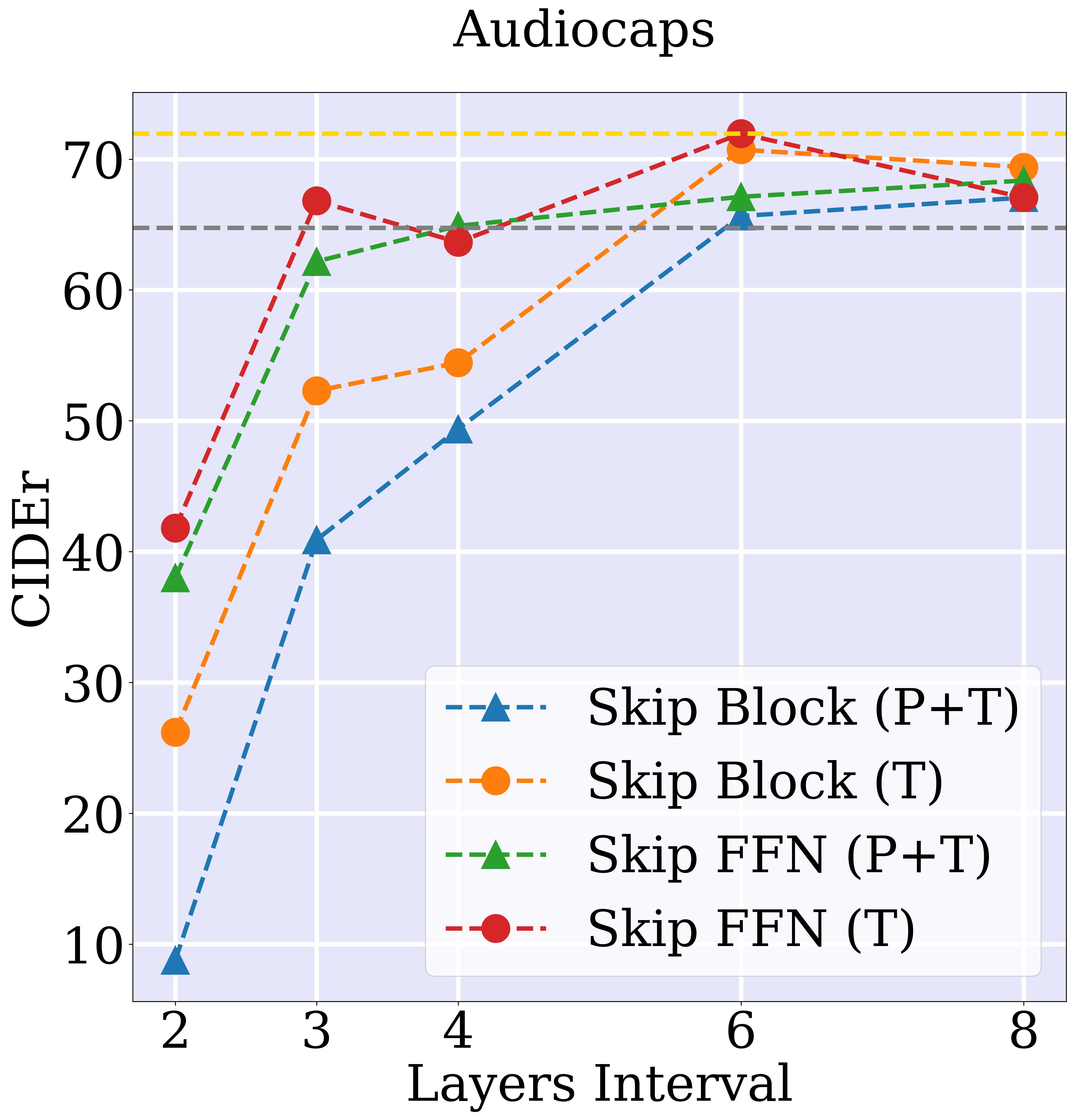}
            \end{subfigure}
        \end{minipage}%

    \end{minipage}%

\caption{\textbf{Which tokens to skip?} We compare between skipping layers, only for the generated textual tokens (T), and all tokens including the prompts (P+T).}
\label{fig:skip_opt_ablation_tokens}
\end{figure*}

\paragraph{Which tokens to skip?} In this comparison, we examine the impact of skipping only the generated textual tokens, as done in the previous section, versus skipping all tokens, including the prompt (P) containing perceptual tokens, the BOS token, and the textual tokens corresponding to questions in QA tasks. As illustrated in \Cref{fig:skip_opt_ablation_tokens}, for QA tasks, we observe that skipping up to 25\% of the blocks for all tokens maintains 90\% of the performance. Generally, skipping layers for the generated tokens yields higher scores.

\begin{figure*}[h]
    \centering
    \begin{minipage}{\linewidth}
    \centering
        \begin{minipage}{.19\linewidth}
        \begin{subfigure}[b]{\textwidth}
                \includegraphics[width=1.0\textwidth]{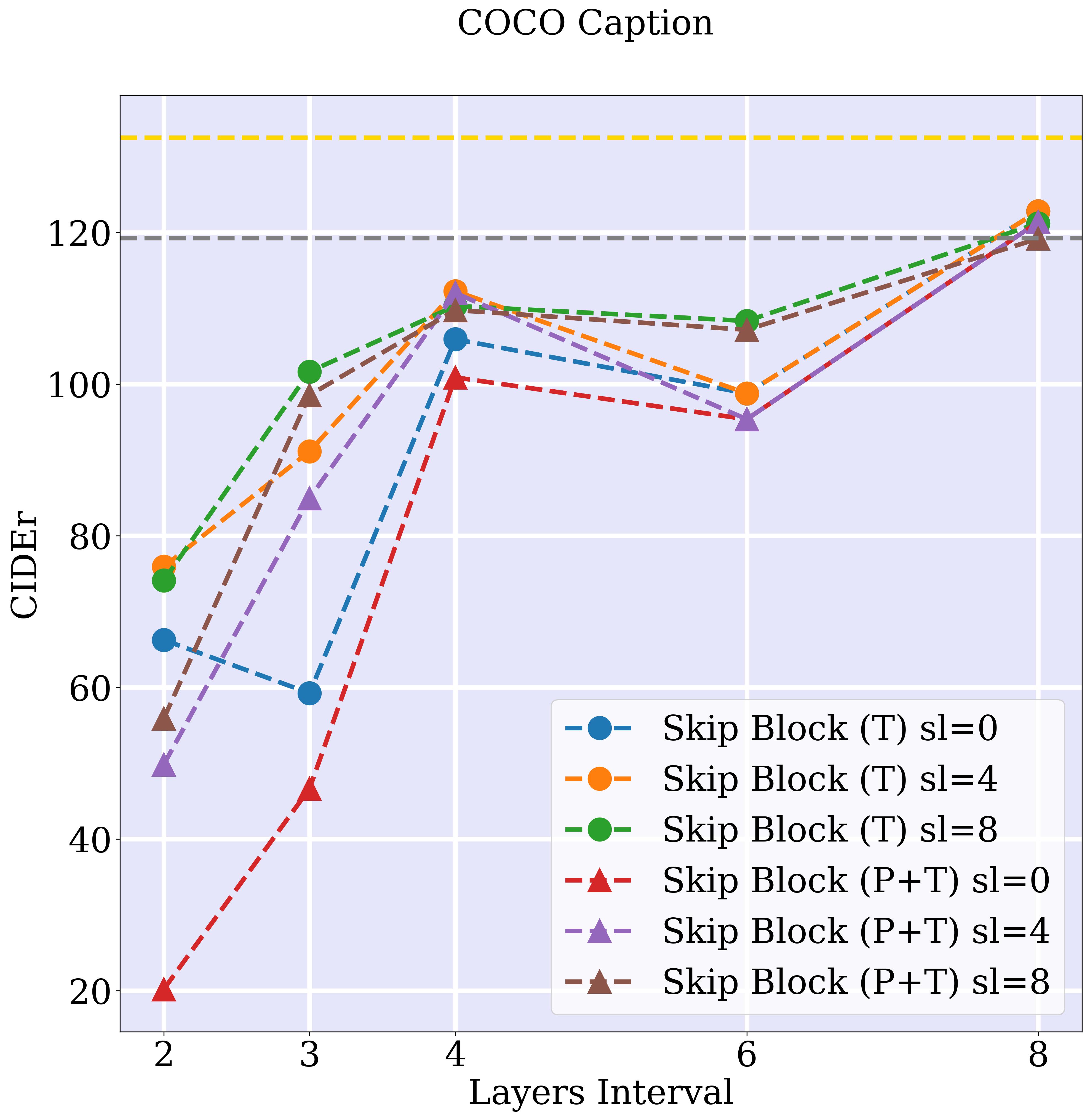}
            \end{subfigure}
        \end{minipage}%
        \begin{minipage}{.19\linewidth}
        \begin{subfigure}[b]{\textwidth}
                \includegraphics[width=1.0\textwidth]{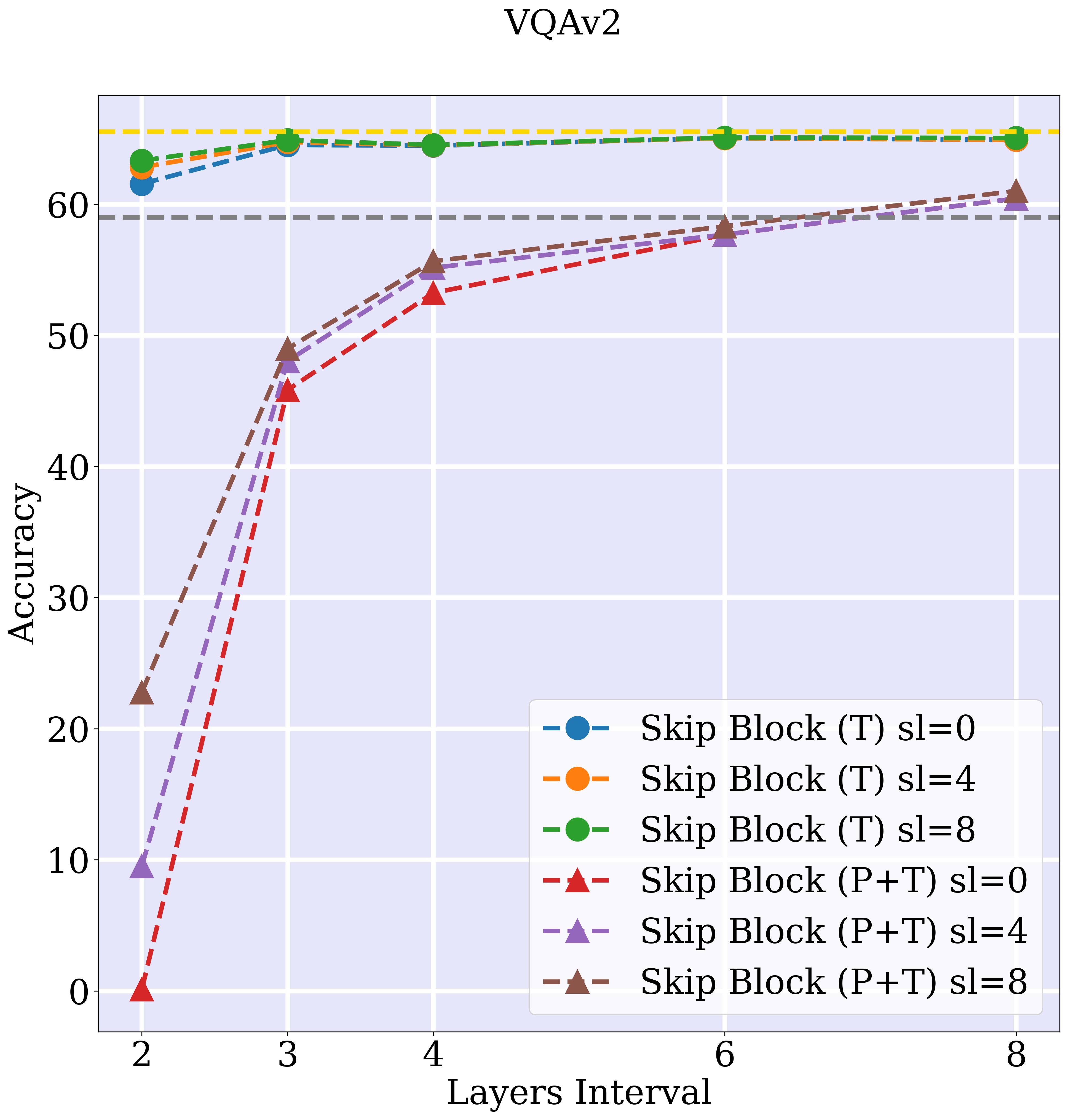}
            \end{subfigure}
        \end{minipage}%
        \begin{minipage}{.19\linewidth}
        \begin{subfigure}[b]{\textwidth}
                \includegraphics[width=1.0\textwidth]{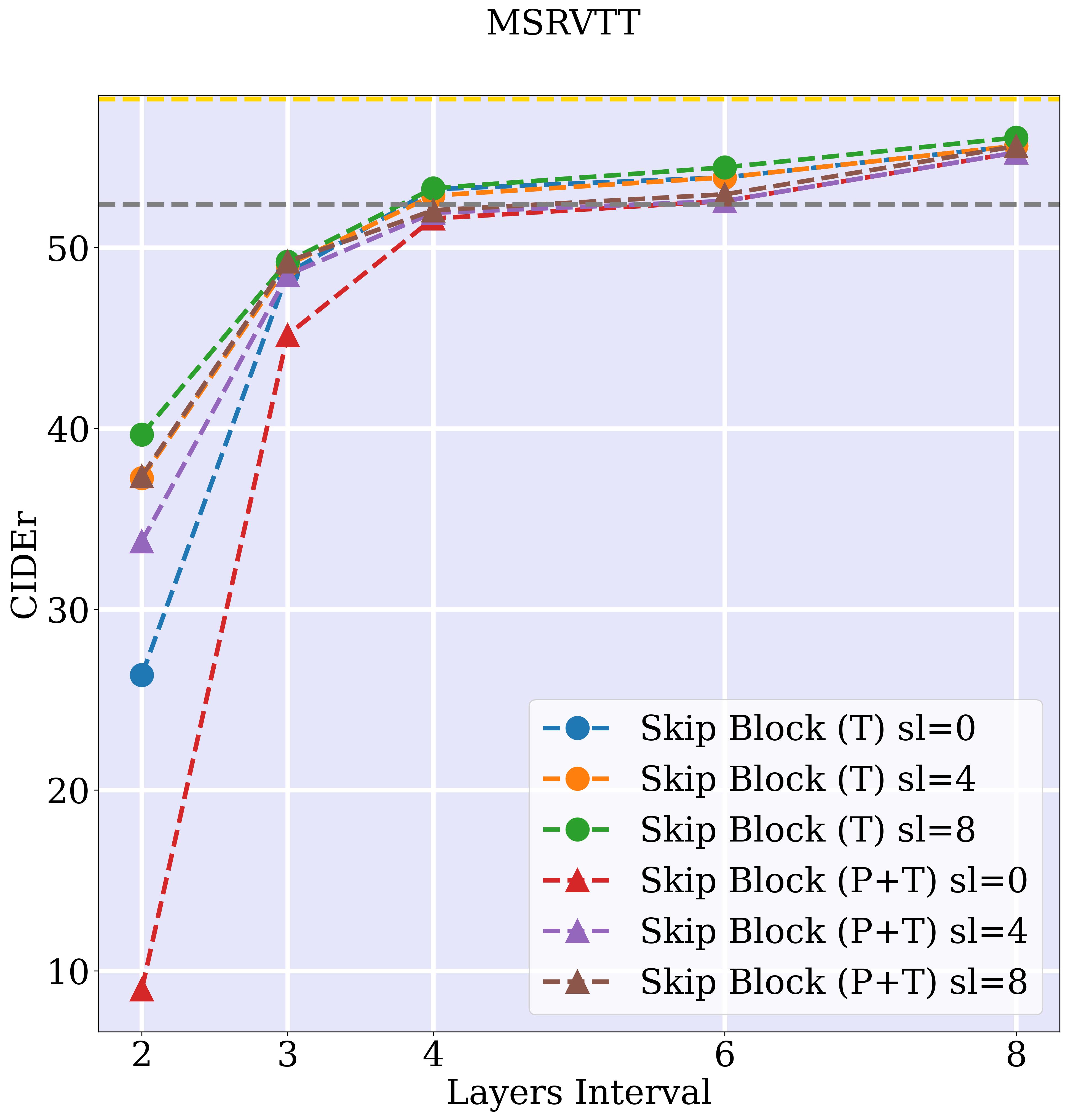}
            \end{subfigure}
        \end{minipage}%
        \begin{minipage}{.19\linewidth}
        \begin{subfigure}[b]{\textwidth}
                \includegraphics[width=1.0\textwidth]{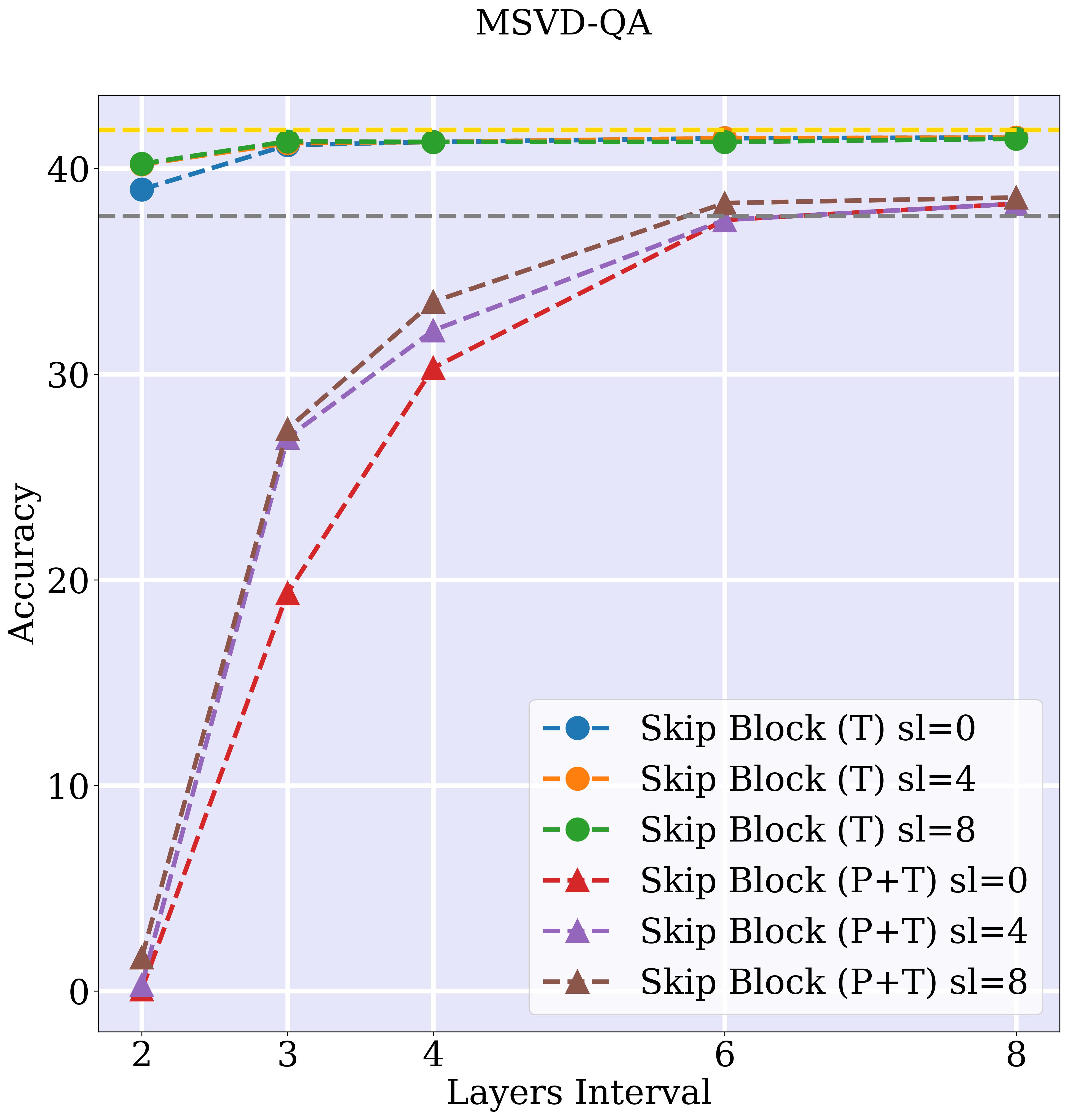}
            \end{subfigure}
        \end{minipage}%
        \begin{minipage}{.19\linewidth}
        \begin{subfigure}[b]{\textwidth}
                \includegraphics[width=1.0\textwidth]{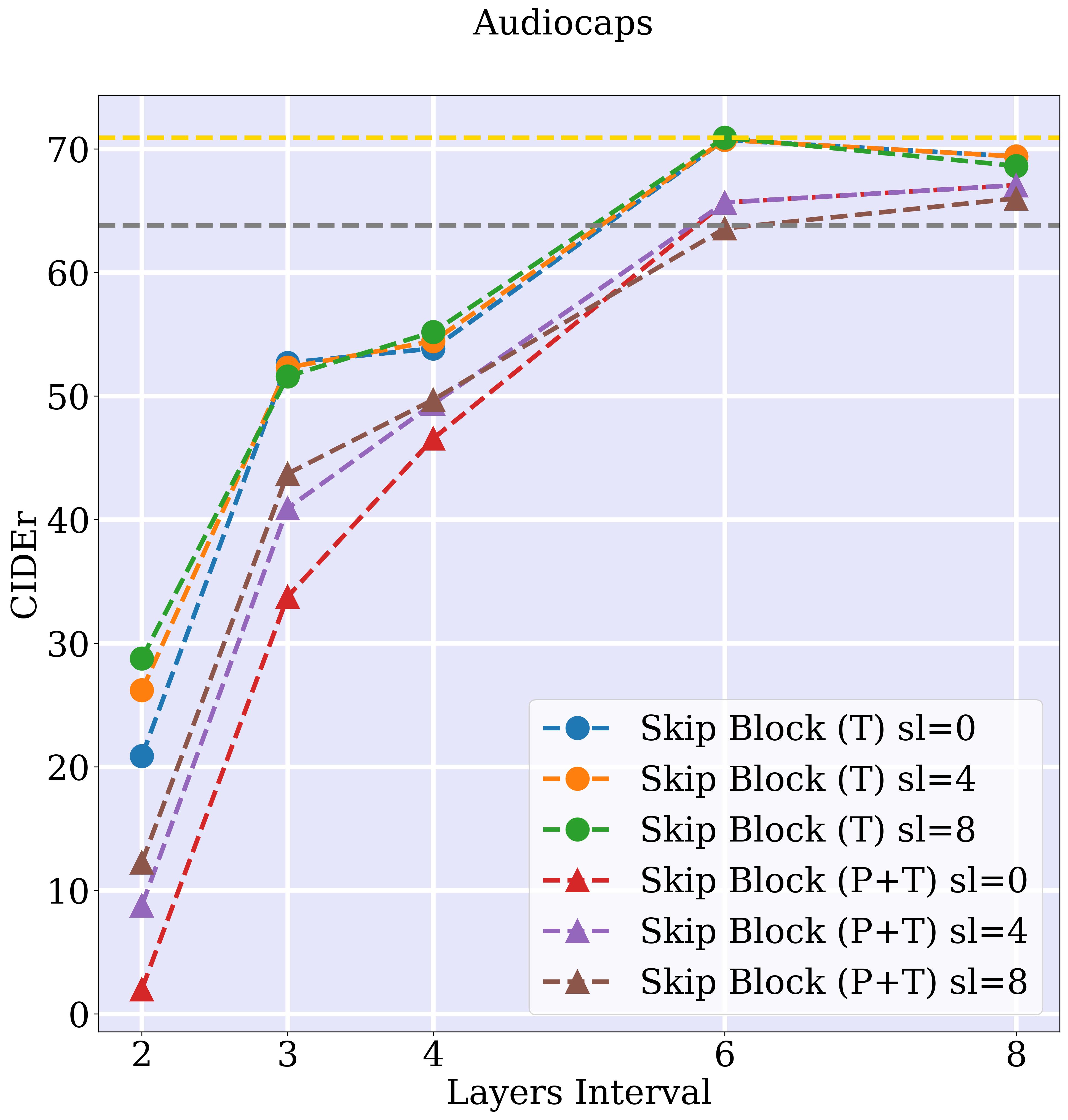}
            \end{subfigure}
        \end{minipage}%

    \end{minipage}%

\caption{\textbf{Where to start skipping layers?} Skipping early layers leads to further decrease in scores. Starting at layer 8 (sl=8) leads to the best performance, especially when skipping many blocks.}
\label{fig:skip_opt_ablation_startlayer}
\end{figure*}

\paragraph{Where to start skipping layers?} 
Prior studies \cite{shukor2024implicit} have highlighted a significant change in both textual and perceptual embeddings at early layers, highlighting their importance, compared to later ones. To further investigate this phenomenon, we compare different starting layers for block skipping. As depicted in \Cref{fig:skip_opt_ablation_startlayer}, we observe that avoiding skipping early layers leads to improvements, particularly when skipping a large number of layers (50\%). This effect is more pronounced when skipping perceptual tokens as well.

\subsection{Parrallelizing computations}

\paragraph{Method.} In this section, we propose to parrallelize different layers, this theoretically helps to reduce the inference time on GPUs by avoiding sequential computations. Specifically, we parrallelize entire blocks by replacing \Cref{eq:block} as follows:
\begin{align}
\label{eq:parrallel_block}
X^{N-1} &= \sum_{\substack{l \geq sl; \ l \ \% \ I = 0; \ l \in \{0, ..., N-2\}}} B^l(X^l) + B^{l+1}(X^{l+1}),
\end{align}
Inside the block, we also parrallelize the FFN and SA each interval (I) of layers by expressing \Cref{eq:in_block} as follows: 
\begin{align}
\label{eq:parrallel_sa}
    X^{l+1} = X^{l} + FC2(g(FC1(LN2(X^{l})))) + SA(LN1(X^{l})) 
\end{align}

\begin{figure*}[h]
    \centering
    \begin{minipage}{\linewidth}
    \centering
        \begin{minipage}{.19\linewidth}
        \begin{subfigure}[b]{\textwidth}
                \includegraphics[width=1.0\textwidth]{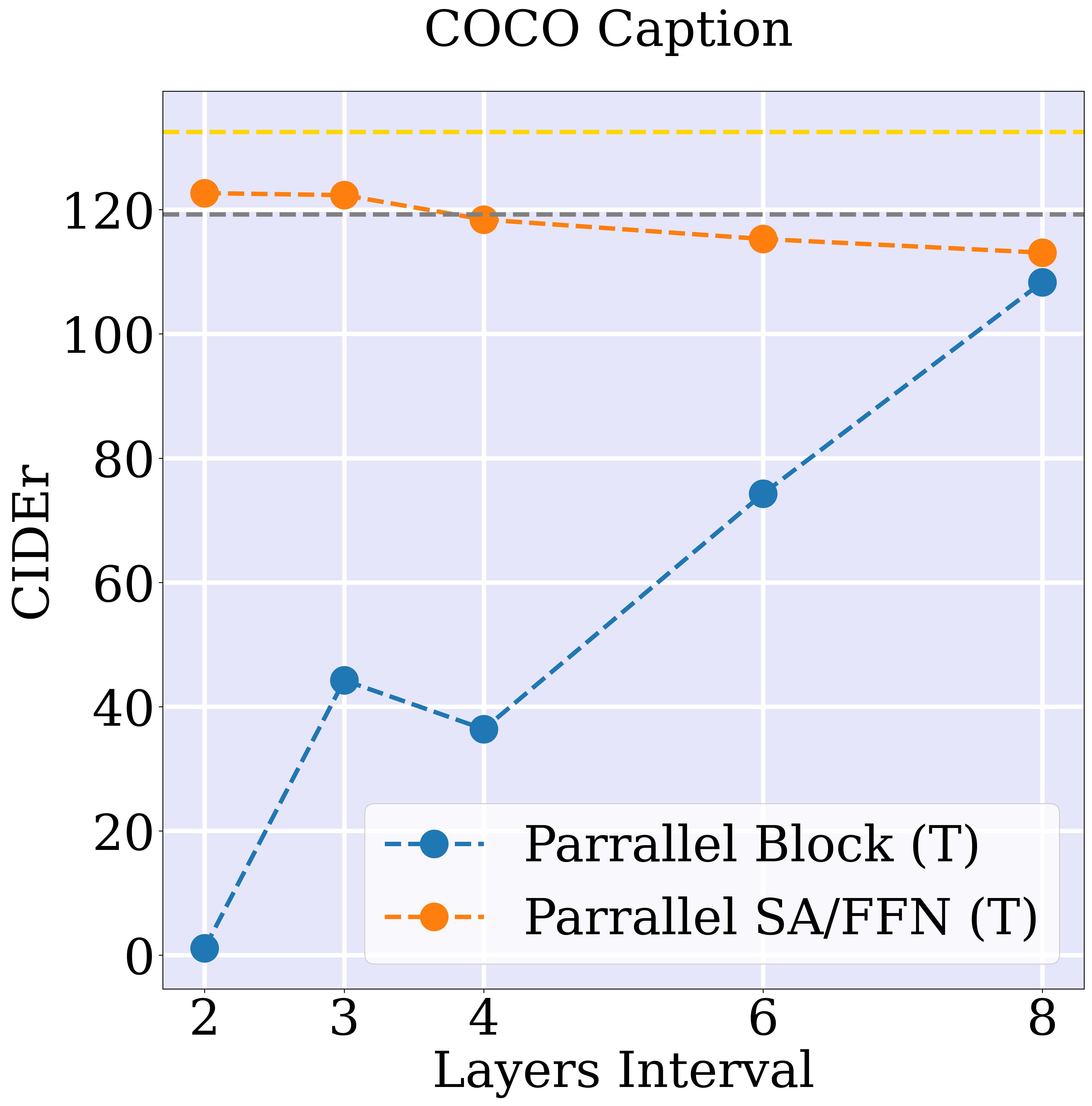}
            \end{subfigure}
        \end{minipage}%
        \begin{minipage}{.19\linewidth}
        \begin{subfigure}[b]{\textwidth}
                \includegraphics[width=1.0\textwidth]{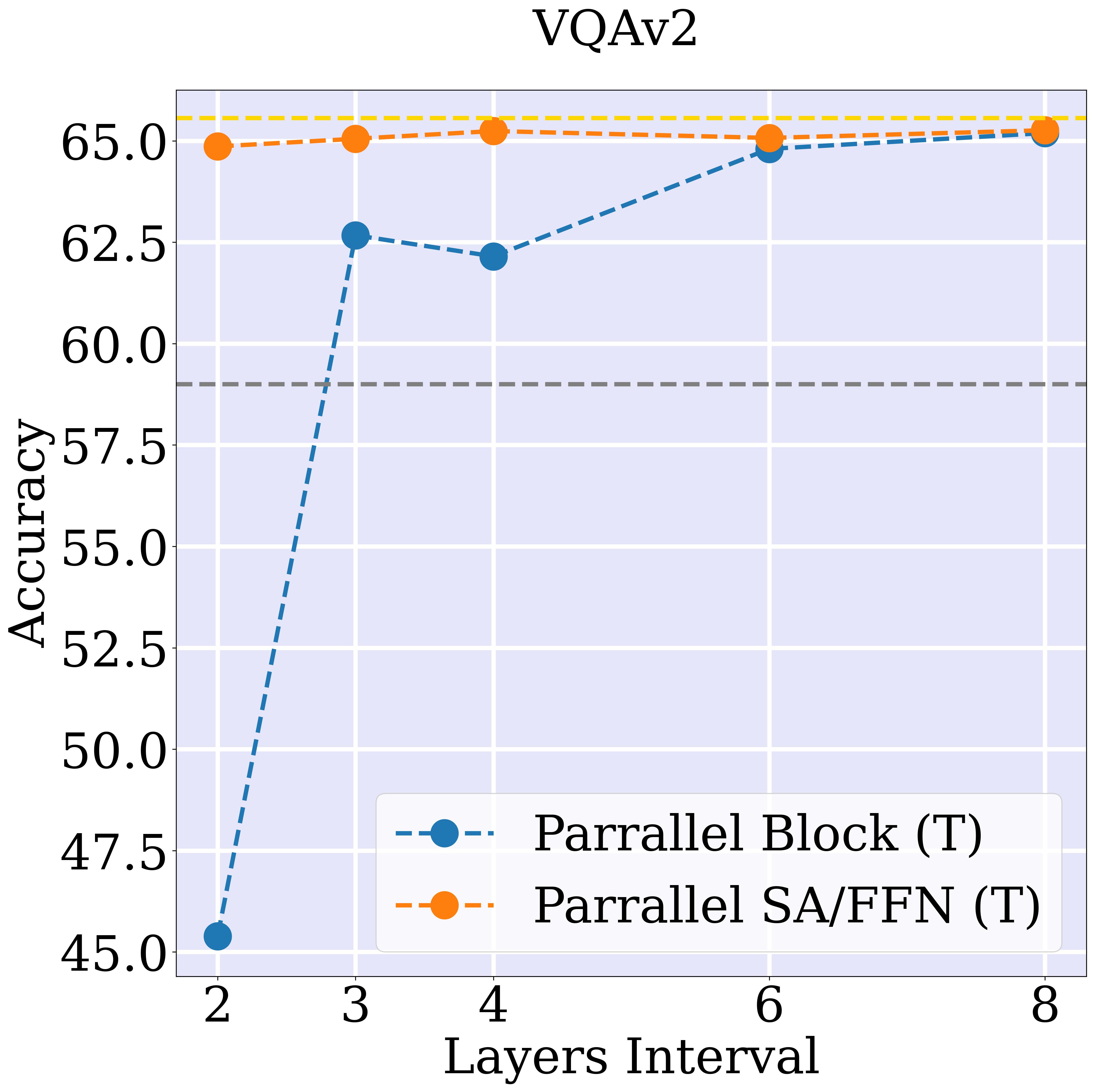}
            \end{subfigure}
        \end{minipage}%
        \begin{minipage}{.19\linewidth}
        \begin{subfigure}[b]{\textwidth}
                \includegraphics[width=1.0\textwidth]{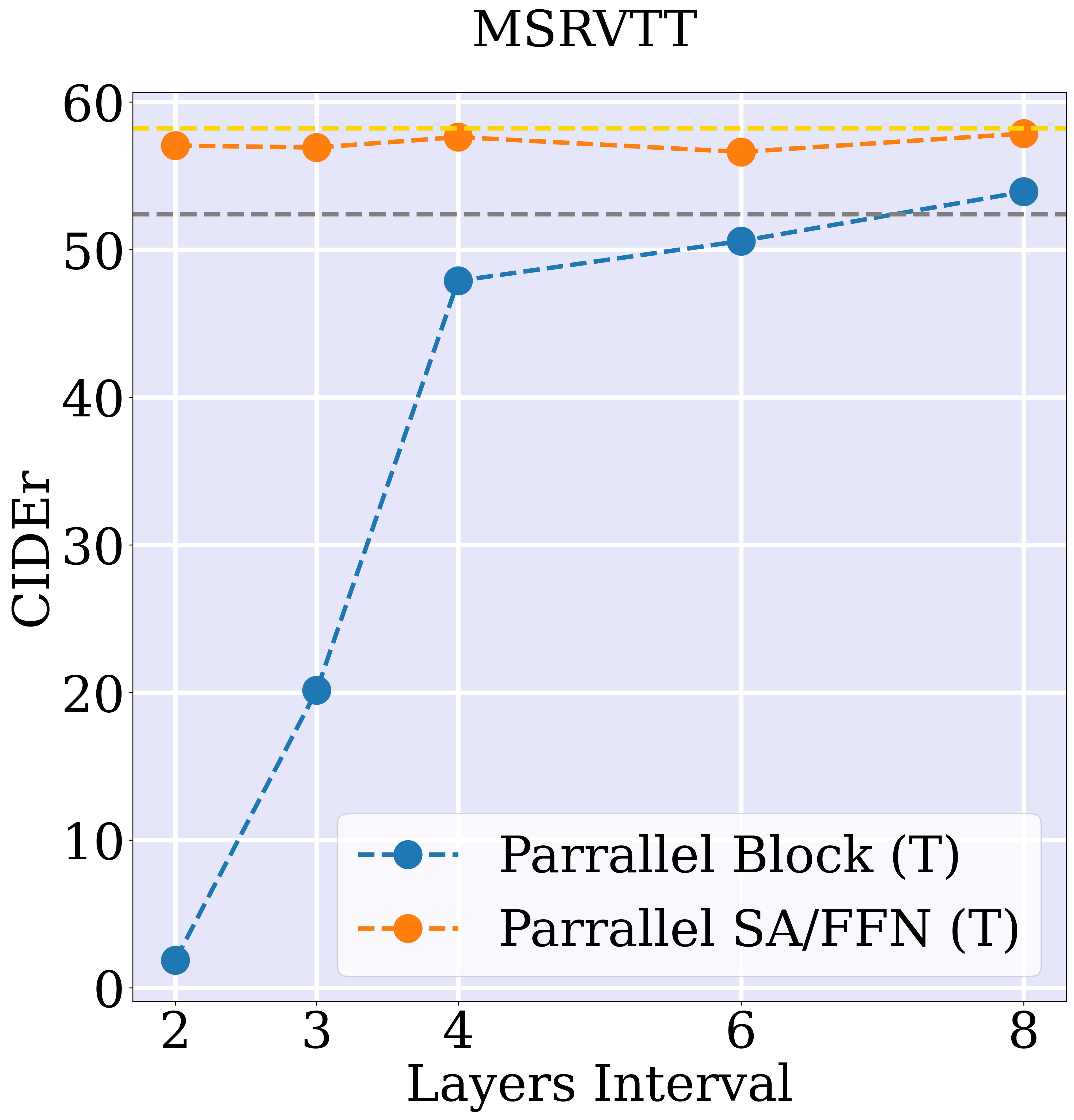}
            \end{subfigure}
        \end{minipage}%
        \begin{minipage}{.19\linewidth}
        \begin{subfigure}[b]{\textwidth}
                \includegraphics[width=1.0\textwidth]{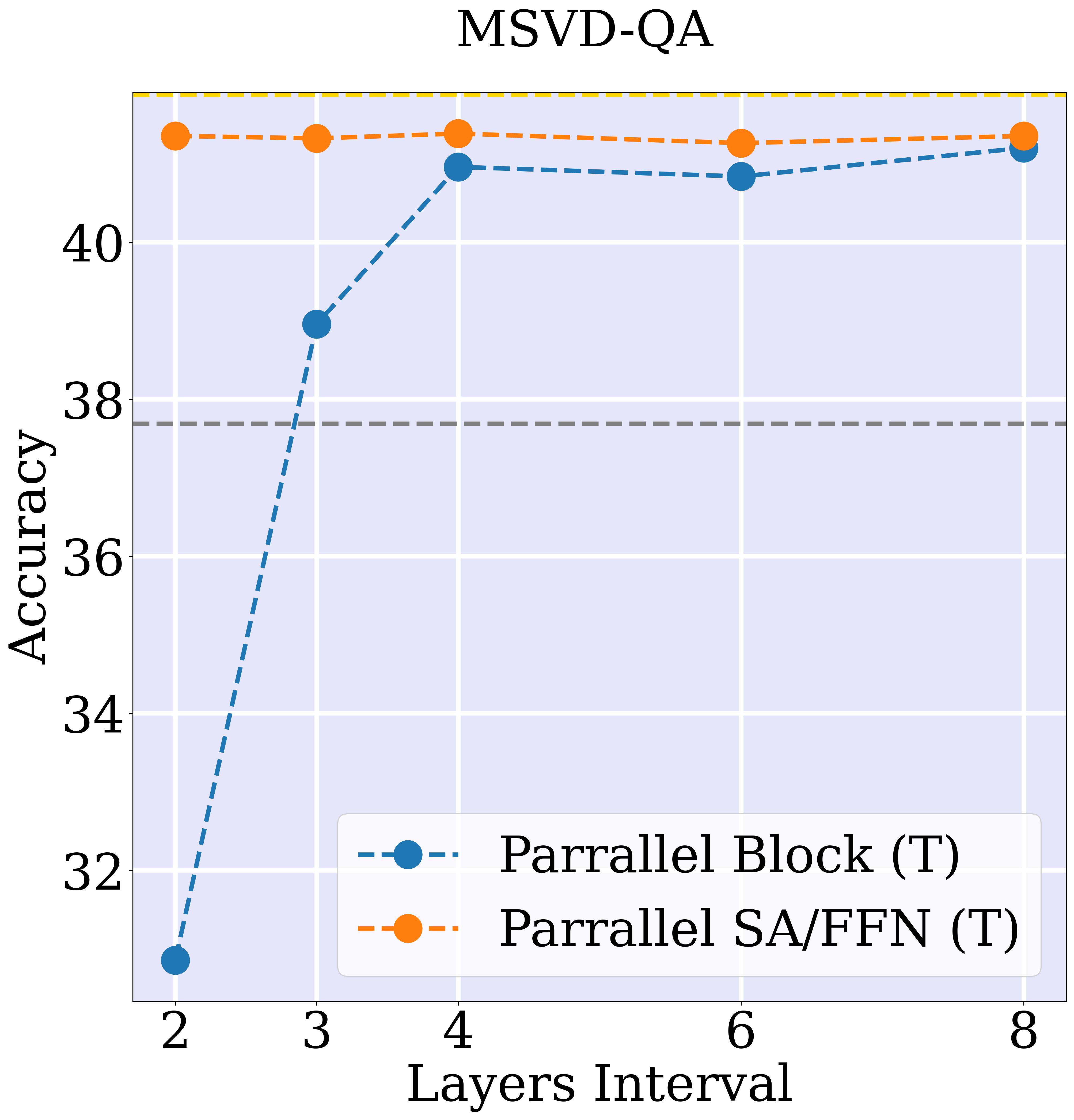}
            \end{subfigure}
        \end{minipage}%
        \begin{minipage}{.19\linewidth}
        \begin{subfigure}[b]{\textwidth}
                \includegraphics[width=1.0\textwidth]{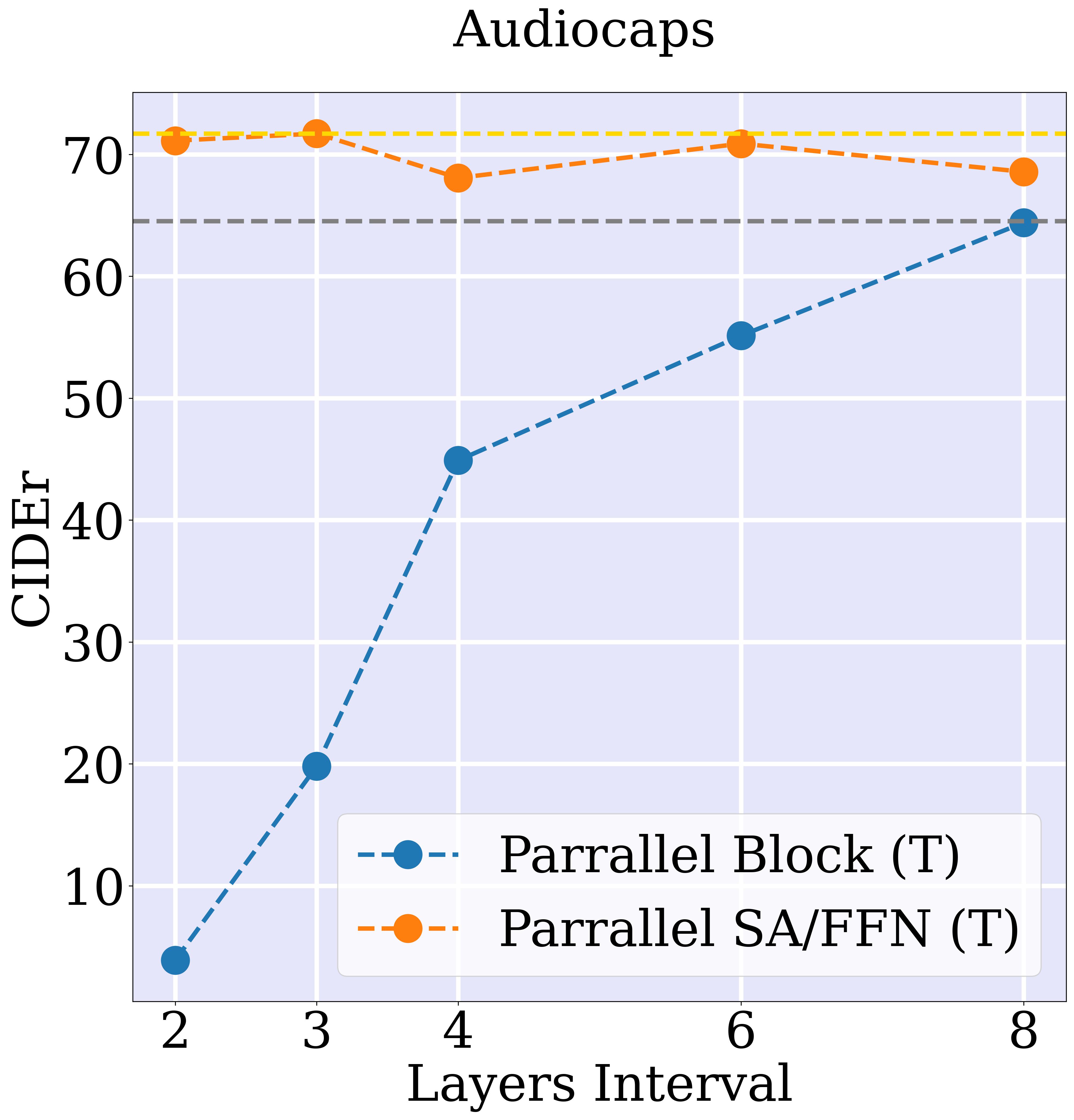}
            \end{subfigure}
        \end{minipage}%

    \end{minipage}%

\caption{\textbf{Parrellelizing computations inside MLLMs.} FFN and SA layers can be cast in parrallel instead of sequential without sacrificing performance. This is less the case for 2 entire blocks.}
\label{fig:parralel_opt_main}
\end{figure*}

\paragraph{Experimental results.} \Cref{fig:parralel_opt_main} compares two approaches to parallelize computations: FFN and SA layers inside each block and parallelizing two entire blocks. The results demonstrate that both approaches perform well for QA tasks. However, parallelizing FFN and SA layers leads to significantly better results on all datasets.

\begin{table}[h]
    \centering
    \small
    \caption{\textbf{Training with highly sparse LLMs}. We train the mapping module from scratch but with compressed LLM.}
    \label{tab:overparam_ablation}
    \resizebox{0.9\linewidth}{!}{%
        \begin{tabular}{lcccccc}
        \toprule	 	
        \multirow{2}{*}{Method}
            & Tr Time
            & \multirow{2}{*}{\#Param}
            & \multirow{2}{*}{\#Tr Param}
            & \multirow{2}{*}{Sparsity}
            & COCO $\uparrow$
            & VQAv2 $\uparrow$
            \\
        \cmidrule(lr{8pt}){6-6}  \cmidrule(lr{8pt}){7-7} 
            & Reduction
            &
            &
            &
            & CIDEr (test)
            & Acc (Val)
            \\

        \midrule
        Baseline 
            & $\times$1 (31 min)
            & 6.7B
            & 7M
            & 0.00 
            & 132.83    %
            & 63.49     %
            \\
        \midrule
        Wanda (task-specific)
            & $\times$1
            & 6.7B
            & 7M
            & 0.50 
            & 126.81    %
            & 55.28     %
            \\
        \midrule
        Random mask 
            & $\times$1
            & 6.7B
            & 7M
            & 0.47 
            & 0.00    %
            & 0.00     %
            \\
        Magnitude (per-out)
            & $\times$1
            & 6.7B
            & 7M
            & 0.50 
            & 19.73    %
            & 2.28     %
            \\
        $\alpha$-SubNet \cite{}
            & $\times$1
            & 6.7B
            & 7M
            & 0.5 
            & 106.77    %
            & 51.77     %
            \\
        \midrule
        Random + Tr 
            & --
            & 6.7B
            & 7M
            & 0.70
            & 111.43    %
            & 33.58   %
            \\
        Magnitude (per-out) + Tr 
            & --
            & 6.7B
            & 7M
            & 0.70
            & 131.37    %
            & 57.67   %
            \\
        Wanda + Tr 
            & --
            & 6.7B
            & 7M
            & 0.50
            & 131.15   %
            & 64.49   %
            \\
        Wanda + Tr 
            & --
            & 6.7B
            & 7M
            & 0.70
            & 131.81    %
            & 58.62   %
            \\
        $\alpha$-SubNet + Tr 
            & --
            & 6.7B
            & 7M
            & 0.70
            & 132.07    %
            & 56.75    %
            \\
        \midrule
        Skip Block (T)  
            & $\times$1
            & 6.7B
            & 7M
            & 0.5$^*$
            & 66.26    %
            & 61.56   %
            \\
        Skip Block (P+T)  
            & $\times$1
            & 6.7B
            & 7M
            & 0.5$^*$
            & 20.21    %
            & 0.13   %
            \\
        Skip Block (P+T)  + Tr 
            & $\times$1.5
            & 6.7B
            & 7M
            & 0.5$^*$
            & 131.53    %
            & 59.13   %
            \\
        \bottomrule
        \end{tabular}
        }
\end{table}

\subsection{Training with highly sparse LLMs}

\paragraph{Training the mapping module from scratch as a remedy for skipping computations.} As seen in previous sections, to preserve the original model performance, the amount of parameters to skip should be limited. In this context, we propose a remedy: training the mapping module from scratch with a compressed LLM. We maintain all hyperparameters and training details as the baseline models, with the only change being the LLM itself, where we either remove weights using Wanda \cite{sun2023wanda} or \alphasubnet \cite{shukor2024implicit}, or skip entire blocks, applicable to all tokens. 

Table \ref{tab:overparam_ablation} presents interesting results. Training with a compressed LLM achieves nearly the same performance for captioning and over 90\% for VQAv2. This holds true when pruning 70\% of the weights or skipping half the blocks (I=2). Moreover, in the case of the latter, besides enhancing inference efficiency, the training time is reduced by a factor of 1.5. For pruning methods, actual efficiency gains necessitate specialized hardware.

\begin{table*}[h]
\small
\centering	
\setlength\tabcolsep{4pt}
\caption{
    \textbf{Comparison between our compressed MLLMs and previous uncompressed ones.} Our models are competitive with previous SoTA despite skipping computations and having smaller number of trainable parameters.
    }
\label{tab:overparam}
\resizebox{0.8\linewidth}{!}{%
\begin{tabular}{lcccccccccccc}
\toprule	 	
\multirow{2}{*}{Method}
    & \multirow{2}{*}{\#P/\#TP/Sparsity}
    & \multirow{2}{*}{Avg}
    & COCO $\uparrow$
    & VQAv2 $\uparrow$
    & MSR-VTT $\uparrow$
    & MSVD-QA $\uparrow$
    & Audiocaps $\uparrow$
    \\
\cmidrule(lr{8pt}){4-4}  \cmidrule(lr{8pt}){5-5} \cmidrule(lr{8pt}){6-6}
\cmidrule(lr{8pt}){7-7}
\cmidrule(lr{8pt}){8-8}
    & 
    &
    & CIDEr (test)
    & Acc (Val)
    & CIDEr (test)
    & Acc (test)
    & CIDEr (test)
    \\
\midrule
ClipCAP \cite{mokady2021clipcap}
    & 7B/3.4M/0.00
    & --
    & 113.08      %
    & --       %
    & --       %
    & --       %
    & --       %
    \\
MAPL \cite{manas2022mapl}
    & 7B/3.4M/0.00
    & --
    & 125.2      %
    & 43.5       %
    & --       %
    & --       %
    & --       %
    \\
eP-ALM~\cite{shukor2023epalm}
    & 6.7B/4M/0.00
    & 63.11
    & 111.6    %
    & 54.9     %
    & 48.79       %
    & 38.40        %
    & 61.86       %
    \\
DePALM~\cite{depalm}
    & 7B/18.1M/0.00
    & --
    & 131.29    %
    & 70.11     %
    & 49.88       %
    & --       %
    & 69.70       %
    \\
\midrule
Baseline 
    & 6.7B/7M/0.00 
    & 72.32
    & 132.83    %
    & 63.49     %
    & 58.23       %
    & 38.83       %
    & 68.24       %
    \\
\midrule
$\alpha$-SubNet \cite{shukor2024implicit}
    & 6.7B/7M/0.47 
    & 50.25  (69.48\%)
    & 106.77    %
    & 51.77     %
    & 38.37       %
    & 31.19       %
    & 23.15       %
    \\
$\alpha$-SubNet + Tr 
    & 6.7B/7M/0.70
    & 70.33  (97.24\%)
    & 132.07    %
    & 56.75    %
    & 57.56      %
    & 38.77       %
    & 66.52       %
    \\
\midrule
Skip Block (P+T)
    & 6.7B/7M/0.5$^*$
    &  6.28  (8.6\%)
    &  20.21   %
    &  0.13   %
    &  8.98     %
    &  0.09      %
    &  2.000      %
    \\
Skip Block (P+T) + Tr 
    & 6.7B/7M/0.5$^*$
    & 71.31  (98.6\%)
    & 131.53     %
    & 59.13     %
    & 58.31       %
    & 39.08        %
    & 68.52        %
\\ 

\bottomrule
\end{tabular}
}
\end{table*}

\paragraph{Comparison with other uncompressed models.} To provide context for our findings, \Cref{tab:overparam} offers a comparison with previous methods utilizing uncompressed LLMs. When training with a sparse LLM, more than 97\% of the overall performance can be retained. Interestingly, despite having almost half the trainable parameters and removing over 50\% of model weights or blocks, our models are competitive with the current state-of-the-art approach (DePALM \cite{depalm}) on nearly all tasks except VQAv2, where we underperform, and MSR-VTT, where we outperform. These results suggest that LLMs are significantly overparametrized for general multimodal tasks.

\begin{figure}[h]
    \centering
    \begin{minipage}{\linewidth}
    \centering
        \begin{minipage}{.49\linewidth}
        \begin{subfigure}[b]{\textwidth}
                \includegraphics[width=1.0\textwidth]{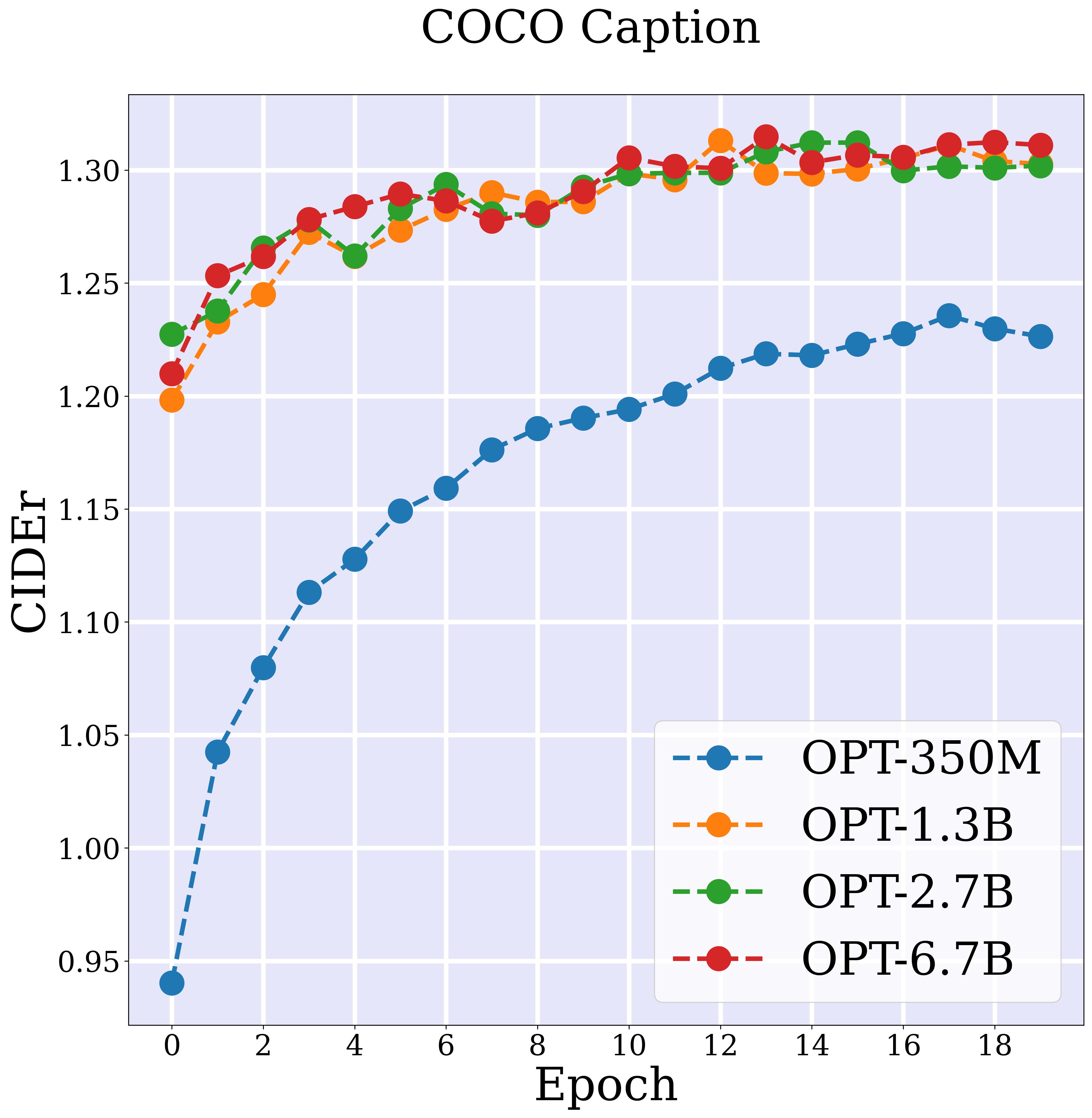}
            \end{subfigure}
        \end{minipage}%
        \begin{minipage}{.49\linewidth}
        \begin{subfigure}[b]{\textwidth}
                \includegraphics[width=1.0\textwidth]{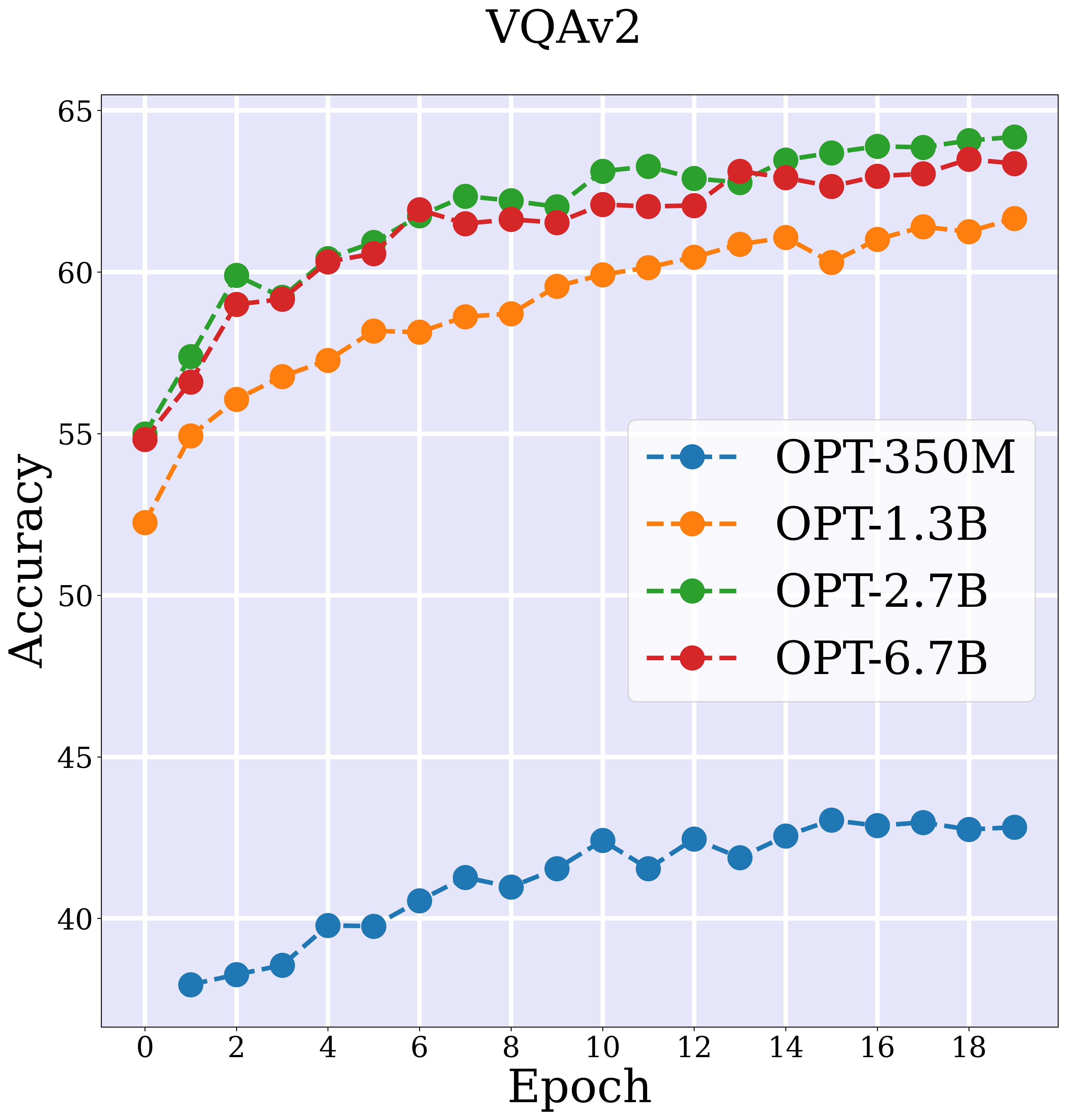}
            \end{subfigure}
        \end{minipage}%
    \end{minipage}%

\caption{\textbf{Training with smaller LLMs.} Training with smaller versions of OPT models leads to comparable performance to larger ones.}
\label{fig:small_llms_ablation}
\end{figure}

\subsection{Training with smaller LLMs}

In previous sections, we explored the feasibility of significantly compressing LLMs for general multimodal tasks. Here, we delve into the possibility of training models with smaller LLMs.

\paragraph{Experimental results.} Our models are trained with smaller versions of the OPT model family. We maintain the same training details as the baseline model, with the exception of the smallest OPT-350M model (where a smaller learning rate yields better results). Figure \ref{fig:small_llms_ablation} presents a comparison between different LLM sizes. Interestingly, models up to 1.3B parameters exhibit the same performance for captioning. Similarly, for VQAv2, we can downscale the model to 2.7B parameters without sacrificing points. However, there is a notable gap, especially for VQAv2, between OPT-350M and other models. Additionally, compared to OPT-350M, larger models are more computationally efficient, as evidenced by the high scores after the first epoch. Table \ref{tab:small_llms_main} further illustrates similar results across additional multimodal tasks. Particularly, the models with OPT-2.7B parameters compete with the baseline and previous approaches using larger LLMs. This suggests the feasibility of training with smaller LLMs, avoiding the high cost associated with larger ones. It is worth noting that previous works \cite{shukor2023epalm} have shown increasing performance with LLM size, but they trained with less powerful visual encoders and for fewer epochs. We argue that proper training of the mapping module (e.g., better encoders, sufficient training) can diminish the improvement coming from larger LLMs.

\begin{table*}[h]
\small
\centering	
\setlength\tabcolsep{4pt}
\caption{
    \textbf{Comparison with previous SoTA when training with smaller LLMs.} Training with OPT-2.7B leads to competitive performance compared to previous SoTA on diverse multimodal tasks.
    }
\label{tab:small_llms_main}
\resizebox{0.8\linewidth}{!}{%
\begin{tabular}{lcccccccccccc}
\toprule	 	
\multirow{2}{*}{Method}
    & \multirow{2}{*}{\#P/\#TP}
    & \multirow{2}{*}{Avg}
    & COCO $\uparrow$
    & VQAv2 $\uparrow$
    & MSR-VTT $\uparrow$
    & MSVD-QA $\uparrow$
    & Audiocaps $\uparrow$
    \\
\cmidrule(lr{8pt}){4-4}  \cmidrule(lr{8pt}){5-5} \cmidrule(lr{8pt}){6-6}
\cmidrule(lr{8pt}){7-7}
\cmidrule(lr{8pt}){8-8}
    & 
    &
    & CIDEr (test)
    & Acc (Val)
    & CIDEr (test)
    & Acc (test)
    & CIDEr (test)
    \\
\midrule
ClipCAP \cite{mokady2021clipcap}
    & 7B/3.4M/0.00
    & --
    & 113.08      %
    & --       %
    & --       %
    & --       %
    & --       %
    \\
MAPL \cite{manas2022mapl}
    & 7B/3.4M
    &
    & 125.2      %
    & 43.5       %
    & --       %
    & --       %
    & --       %
    \\
eP-ALM~\cite{shukor2023epalm}
    & 6.7B/4M
    & 
    & 111.6    %
    & 54.9     %
    & 48.79       %
    & 38.40        %
    & 61.86       %
    \\
DePALM~\cite{depalm}
    & 7B/18.1M
    &
    & 131.29    %
    & 70.11     %
    & 49.88       %
    & --       %
    & 69.70       %
    \\
\midrule
Baseline (OPT-6.7B)
    & 6.7B/7M 
    &  
    & 132.83    %
    & 63.49     %
    & 58.23       %
    & 38.83       %
    & 68.24       %
    \\
\midrule
Baseline (OPT-2.7B)
    & 2.7B/7M 
    &  
    & 132.83    %
    & 64.18     %
    & 57.33       %
    & 40.29      %
    & 65.47       %
    \\
\bottomrule
\end{tabular}
}
\end{table*}

\subsection{Case study for larger scale multitask MLLMs.}
In this section, we investigate the generalization of the proposed approaches to the larger scale multitask setup. We focus on the \llava \cite{liu2023llavaimproved} model which consists of CLIP-ViT-L, Vicuna-v1.5 connected with an MLP and trained on a collection of public datasets. This model is evaluated on more recent multimodal benchmarks such as SEED \cite{li2023seed}, MME \cite{fu2023mme}, POPE \cite{li2023evaluatingpope} ScienceQA (SQA) and \cite{lu2022learnscienceqa}, but also on traditional benchmarks such as VQAv2 \cite{goyal2017makingvqav2}, GQA \cite{hudson2019gqa} and TextVQA \cite{singh2019towardstextvqa}.

\begin{figure}[h]
    \centering
    \begin{minipage}{\linewidth}
    \centering
    \begin{minipage}{.6\linewidth}
        \begin{subfigure}[b]{\textwidth}
                \includegraphics[width=1.0\textwidth]{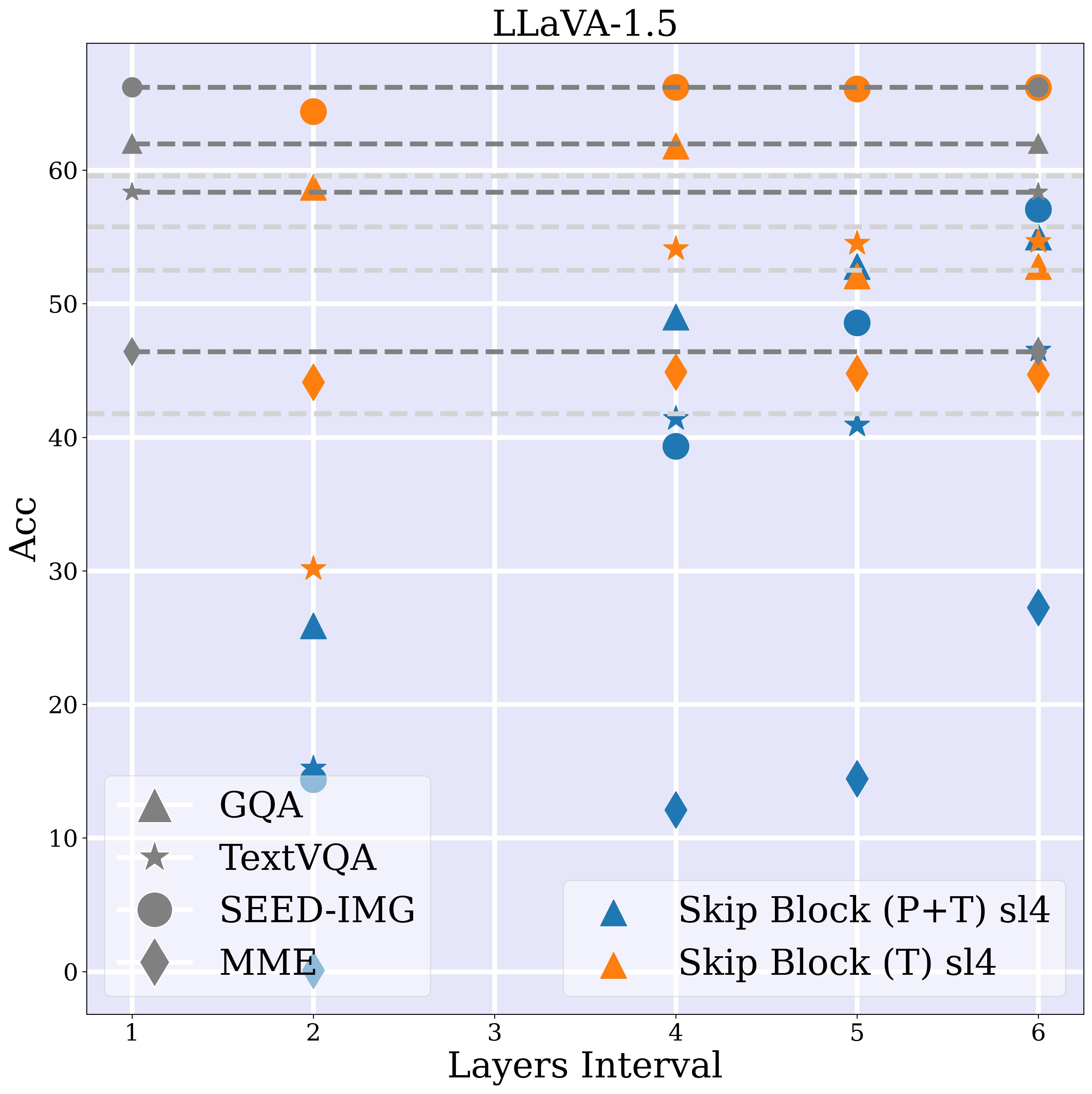}
            \end{subfigure}
        \end{minipage}%
    \end{minipage}%

\caption{\textbf{Skipping computations for \llava.} Skipping entire blocks for textual (T) and all tokens (P+T), each interval of layers. The gray and light gray indicate the original performance and 90\% of it.}
\label{fig:llava_skip}
\end{figure}

\paragraph{Experimental results.} \Cref{tab:llava_skip} shows that we can remove 50\% of the parameters and maintain more than 90\% of the original performance across several datasets. Computing the pruning score based on all tokens gives the best results, followed by the prompt (the contains the visual tokens) and then the textual tokens. We also propose to skip entire blocks (\Cref{fig:llava_skip}). Similar to previous section, skipping only the generated textual tokens leads to the best results where we can retain more than 90\% of the original performance. In general, we notice larger degradation in performance compared to the single task setup (\Cref{sec:skipping_main}). However that results also suggests the possibility of removing redundant computations for large scale models.

\begin{table*}[h]
    \centering
\caption{\textbf{Skipping computations for \llava.} Left: Post-training pruning with Wanda by keeping the weights that are mostly activated by the: prompt (P), textual (T) or all (P+T) tokens.}
\label{tab:llava_skip}
    \begin{minipage}{\linewidth}
    \centering
        \resizebox{\linewidth}{!}{%
        \begin{tabular}{lcccccccc}
        \toprule	 	
        \multirow{2}{*}{Method}
            & \multirow{2}{*}{Sparsity}
            & GQA $\uparrow$
            & VQAv2 $\uparrow$
            & TextVQA $\uparrow$
            & SQA-IMG $\uparrow$
            & COCO $\uparrow$
            & POPE $\uparrow$
            \\
        \cmidrule(lr{8pt}){3-3}
        \cmidrule(lr{8pt}){4-4}  
        \cmidrule(lr{8pt}){5-5} 
        \cmidrule(lr{8pt}){6-6}
        \cmidrule(lr{8pt}){7-7}
        \cmidrule(lr{8pt}){8-8}
            &
            & Acc (test-dev)
            & Acc (test-dev)
            & Acc (Val)
            & Acc (test)
            & CIDEr (test)
            & Acc (test)
            \\
        \midrule
        \llava 
            & 0.0 \%
            &  61.96      %
            &  78.50      %
            &  58.20      %
            &  66.8       %
            &  110.49      %
            &  85.96      %
            \\
        \midrule
        \llava-Wanda (P + T)
            & 50 \%
            & 51.15 (82.58 \%)       %
            & 71.31 (90.82 \%)      %
            & 46.64 (80.06 \%)      %
            & 60.24 (90.23 \%)      %
            & 100.04  (90.52 \%)     %
            & 87.22   (--\%)    %
            \\
        \llava-Wanda (P) 
            & 50 \%
            &  48.1 (77.57 \%)     %
            &  68.76  (87.65 \%)    %
            &  41.94  (72.06 \%)    %
            &  54.44  (81.58 \%)    %
            &  76.99  (69.58 \%)    %
            &  85.37  (99.20 \%)    %
            \\
        \llava-Wanda (T) 
            & 50 \%
            & 43.93 (70.92 \%)      %
            & 64.11 (81.75 \%)     %
            & 41.74 (71.70 \%)      %
            & 51.86 (77.78 \%)      %
            & 94.09 (85.10 \%)     %
            & 78.28 (90.97 \%)      %
            \\
        \bottomrule
        \end{tabular}
        }
        \end{minipage}

\end{table*}

\section{Discussion}

\paragraph{Single-task vs multi-task MLLMs.}  In this work, we focus on parameter and data-efficient models in a single-task setup showing high amount of computation redundancy. We also complement our work with preliminary results on larger-scale models such as \llava that are trained in multitask fashion and can support wider range of tasks, including conducting dialog with humans. On this setup, we also show similar observations, however we think these baselines requires more adapted compression techniques to limit the performance degradation.

\paragraph{Dynamic compute.} In our study, we focus on static computation skipping techniques, where the skipping strategy remains constant regardless of the task or input example. These static approaches are hardware-agnostic and compatible with scaling techniques. However, an alternative direction involves exploring more adaptive compute strategies, which ideally allocate varying amounts of computations based on the task's complexity. While similar approaches have been proposed for LLMs \cite{ainslie2023colt5,schuster2022confident,lei2024conditional,bengio2015conditional,shazeer2016outrageouslymoe,raposo2024mixturemod}, we believe there is still significant room for improvement in this area. Ultimately, we view our study as an initial step in highlighting the overparameterization of LLMs and advocating for greater efforts to reduce their computational costs.

\paragraph{Limitations.} Our study primarily focuses on parameter-efficient MLLMs, where the LLM remains frozen. We acknowledge that there are other architectures we did not explore, such as those involving interaction through cross-attention mechanisms \cite{alayrac2022flamingo,laurencon2023obelics}. Besides, we did not delve into more complex multimodal tasks that necessitate reasoning capabilities \cite{yue2023mmmu}. Because the main objective of the paper does not include proposing new SoTA model compression technique, we did not extensively compare with more recent approaches. Extending our study to encompass these scenarios represents an important avenue for future research. In addition, we focus on investigating the redundancy that could lead theoretically to high efficient training and inference. However, the reduction is expected to be smaller at actual devices as there are many other affecting factors.

\section{Conclusion}

This study investigates the redundancy of computations in perceptually augmented LLMs (MLLMs) across various granularity levels. Our experiments reveal the potential for significant reduction in computations by skipping entire blocks, FFN layers, and even individual neurons. We demonstrate that training the mapping module with severely compressed LLMs can effectively preserve over 97\% of performance. Alternatively, training with smaller LLMs can achieve comparable performance to models two or three times larger. We show similar findings across both single-task and multitask multimodal settings, underscoring their broad applicability. We hope that this work will encourage future works to focus on methods to reduce the computation cost of MLLMs at both training and inference stages.

\section{Acknowledgments} 
The authors would like to thank Arnaud Dapogny and Edouard Yvinec for fruitful discussions, and Damien Teney and Alexandre Ramé for their helpful feedback on the paper. This work was partly supported by ANR grant VISA DEEP (ANR-20-CHIA-0022), and HPC resources of IDRIS under the allocation 2024-[AD011013415R2] made by GENCI.

\bibliographystyle{splncs04}
\bibliography{main}

\end{document}